\RequirePackage{fix-cm}
\documentclass[twocolumn]{svjour3}          
\smartqed  

\usepackage{cite}
\usepackage{amsmath,amssymb,amsfonts}
\usepackage{algorithmic}
\usepackage{makecell}
\usepackage{graphicx}
\usepackage{textcomp}
\usepackage{array} 
\usepackage{longtable}
\usepackage{booktabs}
\usepackage{float}
\journalname{Archives of Computational Methods in Engineering}
\begin{document}

\title{A Comprehensive Review of Markov Random Field and Conditional Random Field Approaches in Pathology Image Analysis
}

\titlerunning{Yixin Li et al.}        

\author{Yixin Li\and Chen Li (Corresponding author) \and Xiaoyan Li (Corresponding author) \and Kai Wang\and Md Mamunur Rahaman\and Changhao Sun\and Hao Chen\and Xinran Wu\and Hong Zhang \and Qian Wang}


\institute{
Yixin Li \at
Microscopic Image and Medical Image Analysis Group, MBIE College, Northeastern University, Shenyang, China \\
\email{1047792668@qq.com}           
\and
Chen Li \at
Microscopic Image and Medical Image Analysis Group, MBIE College, Northeastern University, Shenyang, China \\
\email{lichen201096@hotmail.com}           
\and                
Xiaoyan Li \at
Cancer Hospital of China Medical University, Shengyan, China \\
\email{lixiaoyan@cancerhosp-ln-cmu.com}   
\and
Kai Wang \at 
Shenyang Institute of Automation, Chinese Academy of Sciences, Shenyang, China \\
\email{wangkai@sia.cn}  
\and
Md Mamunur Rahaman \at
Microscopic Image and Medical Image Analysis Group, MBIE College, Northeastern University, Shenyang, China \\
\email{mamunrobi35@gmail.com}           
\and
Changhao Sun \at
Microscopic Image and Medical Image Analysis Group, MBIE College, Northeastern University, Shenyang, China \\
\email{sch236@sina.com}           
\and
Hao Chen \at
School of Nanjing University of Science and Technology, Nanjing, China \\
\email{451091574@qq.com}  
\and
Xinran Wu \at
Microscopic Image and Medical Image Analysis Group, MBIE College, Northeastern University, Shenyang, China  \\
\email{975990071@qq.com} 
\and
Hong Zhang \at 
Shengjing Hospital of China Medical University, Shenyang, China    \\
\email{haojiubujian1203@sina.cn}        
\and   
Qian Wang \at
Cancer Hospital of China Medical University, Shengyan, China      \\
\email{wangqian$\_$an$\_$16@163.com} \\
*This article has been published in Archives of Computational Methods in Engineering on 27 April 2021.         
}

\date{Received: date / Accepted: date}

\maketitle

\begin{abstract}
Pathology image analysis is an essential procedure for clinical diagnosis of numerous diseases. 
To boost the accuracy and objectivity of the diagnosis, nowadays, an increasing number of intelligent systems are proposed. Among these methods, random field models play an indispensable role in improving the investigation performance. In this review, we present a comprehensive overview of pathology image analysis based on the \emph{Markov Random Fields} (MRFs) and \emph{Conditional Random Fields} (CRFs), which are two popular random field models. First of all, we introduce the framework of two random field models along with pathology images. Secondly, we summarize their analytical operation principle and optimization methods. Then, a thorough review of the recent articles based on MRFs and CRFs in the field of pathology is presented. 
Finally, we investigate the most commonly used methodologies from the related works and discuss the method migration in computer vision.

\end{abstract}

\section{Introduction}
\label{sec:introduction}

\subsection{Markov Random Fields}
\textit{Markov Random Fields} (MRFs), one of the classical undirected \textit{Probabilistic 
Graphical Models}, belongs to the Bayesian framework~\cite{grenander1983tutorials}. Each object of the MRFs, which is called a ‘node’ (or vertex or point) in the graphical models, denotes a random variable and connected by an ‘edge’ between them. In image analysis domain, the MRFs can describe the pixel-spatial interaction due to its structure and thus is developed initially to analyze the spatial relationship of physical phenomena. However, the number of possible states of MRFs is excessively broad, and its 
joint distribution is hard to be calculated~\cite{monaco2012class}.

In 1971, Hammersley and Clifford proved the equivalence between the MRFs and Gibbs 
distribution and this theory is lately developed by~\cite{besag1974spatial}. An 
explicit and elegant formula models the joint distribution of MRF due to the MRFs-Gibbs 
equivalence, where the probability is described by a potential function~\cite{li1995MRF}. 
The proposed optimization algorithms such as Iterative Condition Mode (ICM) and Expectation 
Maximization (EM) and MRFs-Gibbs equivalence make the MRFs a practical model. Additionally, 
as mentioned above, the MRF model considers the important spatial constraint, which is 
essential to interpret the visual information. Hence, the MRFs have attracted significant 
attention from scholars since it is proposed.

The very first research performed to segment medical images using MRFs are 
in~\cite{zhang2001segmentation}. After persistent and in-depth research in last 
two decades, the MRFs are now widely adopted to solve problems in computer vision, consisting of image reconstruction, image 
segmentation as well as image classification~\cite{wang2012gmm}. 
However, MRFs also have limitations: the joint probability of the images and its annotations are modelled due to the underlying generative nature of MRFs, resulting in a complicated structure that requires a large amount of computation.
 The high difficulty of parameter estimation due to 
the huge amount of parameters based on the MRF becomes the bottleneck of its 
growth~\cite{wu2018image}.

\subsection{Conditional Random Fields}
\textit{Conditional Random Fields} (CRFs), as an important and prevalent type of 
machine learning method, is constructed for data labeling and segmentation. Contrary to generative nature of MRF,it is an undirected discriminative graphical model 
focusing on the posterior distribution of observation and possible label sequence 
~\cite{yu2019comprehensive}.
Developed based on the Maximum Entropy Markov Models (MEMMs)~\cite{mccallum2000maximum}, the CRFs avoid the fundamental limitations and of it and other directed graphical models like Hidden Markov Model 
(HMM)~\cite{rabiner1986introduction}.

Compared to the Bayesian models proposed before, the CRF models have three main advantages: 
First, the CRF models solve the label bias problem of MEMMs, which is the main deficiency. 
Second, the CRFs models the probability of a label sequence for a known observation array, which is called condition probability.
Compared to other generative models whose
usual training goal for the parameters is  the joint probability function maximization. In contrast, the CRF models are not required to traverse all possible observation sequences, which is typically intractable. Thirdly, the CRFs relax the 
strong dependencies assumption in other Bayesian models based on directed graphical models 
and are capable of building higher-order dependencies, which means that the results of CRFs 
are more closer to the true distribution of the data. The CRF models have many different 
variants, such as Fully-connected CRF (FC-CRF) and deep Gaussian CRF. The Maximum A Posteriori Estimation (MAP) method is usually utilized for unknown parameters inference of the CRFs~\cite{lafferty2001conditional}.

The CRFs attract researchers' interest in the domain of machine learning because various application scenarios can apply CRFs achieving better results, such as for Name Entity Recognition Problem in Natural Language 
Processing~\cite{zhang2017semi}, Information Mining~\cite{wicaksono2013toward}, Behavior Analysis~\cite{zhuowen2013human}, Image and Computer Vision~\cite{kruthiventi2015crowd}, 
and Biomedicine~\cite{liliana2017review}. In recent year, with the rapid development of 
deep learning (DL), the CRF models are usually utilized as an essential pipeline within the 
deep neural network in order to refine the image segmentation results~\cite{wu2018image}. 
Some researches~\cite{zhuowen2013human} incorporate them into the network architecture, 
while others~\cite{chen2014semantic} include them into the post-processing step. Studies 
show that they mostly achieve better performances than before.

\subsection{Pathology Image Analysis}
The term ‘pathology’ has different meanings under different conditions. It usually refers 
to histopathology and cytopathology, and under other circumstances, it refers to other 
subdiscipline. To avoid confusion, the term ‘pathology’ refers to histopathology and 
cytopathology in this paper, which detect morphological changes of a lesion like tissue and 
cellular structure under a microscope. For purpose of obtaining a tumor sample, performing a biopsy or aspiration requiring intervention such as an image-guided procedure 
or endoscopy is a real necessity~\cite{world2019guide}.

Pathology image analysis is considered as one of the key elements of early diagnosis of various diseases, especially cancer, reported by the World Health Organization (WHO)~\cite{world2017guide}. Cancer is now responsible for nearly $16.7\%$ death in 
the world, and over 14 million people are diagnosed as cancer every year. It is found 
that the ability to provide screening and early cancer diagnosis has a significant 
impact on improving the curing rate, reducing mortality over cancer and cutting treatment 
costs, because the main problem now is that numerous cancer cases are already at an advanced stage when they are diagnosed~\cite{WHO2017Early}. An accurate pathologic diagnosis is critical, and the 
reasons are as follows: First, the definitive diagnosis of cancer and other diseases 
like retinopathy must be made by morphological and phenotypical examination of suspected 
lesion. Second, especially for cancer, pathology image analysis is essential to determine 
the degree of tumor spread from the original site, which is also called staging. Thirdly, 
the appropriate treatment afterwards relies on various pathologic parameters, including 
the type, grade and extent of cancer~\cite{world2019guide,WHO2020cancer}.

Traditionally, pathology results are provided by manual assessment. However, due to 
the laborious and tedious nature of pathologists’ work as well as the complexity and 
heterogeneity, the manual analysis is relatively subjective and even leads to 
misdiagnosis~\cite{janowczyk2016deep}. With the rapid advance of technology, 
Computer-Aided Diagnosis (CAD) emerges, which promises hopefully more standardized 
and objective diagnosis comparing to manual inspection. Besides, CAD can offer 
quantitative result. In the preceding decades, there have been numerous researches in CAD area, proposing various algorithms combining prior knowledge and training data in order to help pathologists make clinical diagnosis and researchers in studying disease mechanisms~\cite{fuchs2011computational}. 

\subsection{Paper Searching and Screening}
\begin{figure}[htbp]
  \centering
  \centerline{\includegraphics[width=0.98\linewidth]{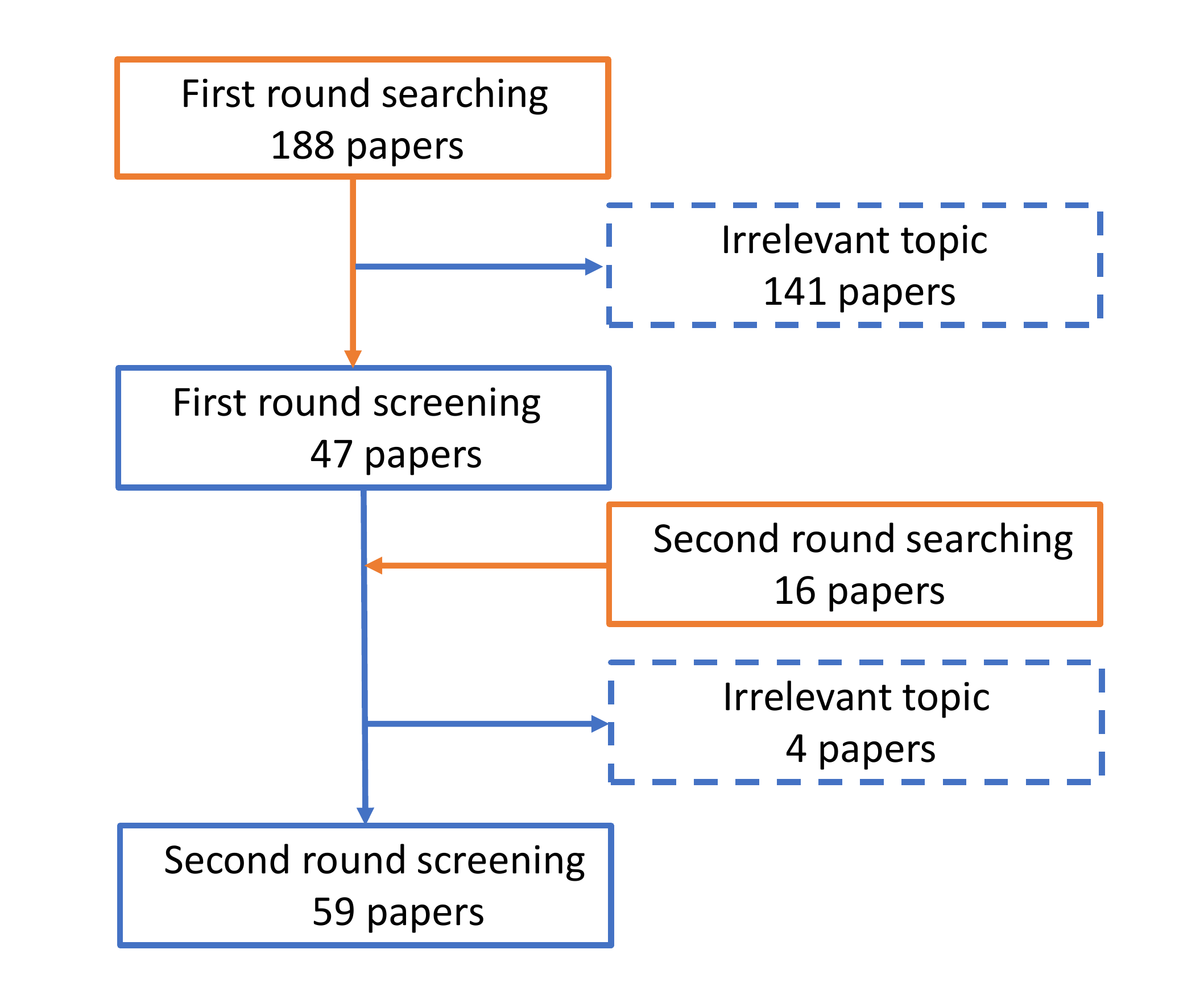}}
\caption{The flowchart of the paper searching and screening process.}
\label{fig:app}
\end{figure}

To collect the papers related to our research interest, we carry out a series of paper searching and screening with Google Scholar's help. The whole process is presented in Fig.~\ref{fig:app}. During the first round of paper searching, the keyword is applied for a wide range of searches, and a total of 188 papers are found. After carefully reading the abstract, 141 papers related to the imaging principle or other imaging methods are precluded. And 47 papers are retained, including 28 papers related to microscopic image analysis, 11 papers related to micro-alike images analysis, and ten review papers. Afterward, the second research mainly focuses on the reference in the above 47 papers and the review, and 16 papers are selected from a large number of citations. Finally, 59 research papers remain and to be concluded in our review. 

\subsection{Motivation of This Review}
Pathology image analysis influences disease detection in a significant way. However, 
manual detection of microscopic images requires exhaustive examination and analysis 
by a small number of experienced pathologists, and the detection results may be 
subjective and vary from different pathologists. Especially for Whole-slide Images (WSIs), 
which are extensively large (e.g., $100,000\times200,000$ pixels), it is labor-intensive 
and time-consuming to achieve fine-grained results~\cite{kong2017cancer}. Therefore, 
various intelligent diagnosis systems are introduced to assist pathologists in detecting 
diseases. In this process, image labeling serving as the intermediate step is in a 
significant position. To some degree, the accuracy of the whole process is associated 
with the quality of the image labeling. In terms of labeling, each pixel’s label is not 
only related to its individual information but also depends on its 
neighborhood~\cite{wu2018image}. For instance, a WSI is divided into small patches, 
when a patch is labeled as a tumor, there is a probability that its neighbouring patches 
are also labeled as tumor~\cite{kong2017cancer}. Moreover, researches are suggesting 
that the distribution of labels in pathology images has a specific underlying structure 
proved to be beneficial to diagnosis~\cite{doyle2008automated}. However, some existing 
methods do not take the contextual information on neighboring labels into consideration. 
For example, some traditional binary classifiers like Support Vector Machine (SVM) and 
Maximum Entropy only consider one single input and ignore the spatial relationship with 
other inputs while predicting the labels~\cite{yu2019comprehensive}. Besides, the 
advanced DL models also have this problem. Although the Convolutional Neural Network (CNN) 
has a large number of input images, the spatial dependencies on patches are usually 
neglected, and the inference is only based on the appearance of individual 
patches~\cite{zanjani2018cancer}. Hence, the structural model is proposed to solve 
this problem. The most prevalent models are the MRFs and the CRFs, which explicitly 
model the correlation of the pixels or the patches being 
predicted~\cite{arnab2018conditional}. A better result can be obtained if the 
information from the neighbouring patches are integrated into the MRFs or the CRFs. 
Therefore, by incorporating them into the CNNs, the small spurious regions like noisy 
isolated predictions in the original output are almost eliminated~\cite{fu2016deepvessel}. 
Meanwhile, the boundaries are proved to be refined and become 
smoother~\cite{manivannan2014brain}.

According to our survey, a limited number of reviews concentrate on medical image analysis and random fields.
Among these studies, \cite{yu2019comprehensive} focuses on the CRFs and their application for different area. Some 
surveys~\cite{wang2013markov,wu2018image} concentrate on image analysis with random 
field models. However, those papers rarely refer to medical image analysis, let alone 
pathology image analysis. Additional papers direct on Artificial Intelligence in 
pathology image analysis~\cite{chang2019artificial,li2020review}, such as using 
the DL algorithm~\cite{litjens2017survey,wang2019pathology,rahaman2020survey}
and other image analysis techniques~\cite{he2012histology,xing2016robust}. Those papers 
introduce various algorithms or models, but the literature quantity about random field 
models is too limited to discuss them specifically, which is inconsistent with their 
importance. In the following paragraphs, eleven of these reviews are listed and 
analyzed in detail.

He et \textit{al}.~\cite{he2012histology} publishes a research survey in 2012, presenting a summary of CAD techniques in histopathology image analysis domain, aiming at detecting and classifying carcinoma automatically. This paper also introduces the MRFs in a separate paragraph. However, this paper refers to around 158 related works, only including seven papers related to the MRFs or CRFs.

Wang et \textit{al}.~\cite{wang2013markov} in 2013 provides an overview of MRFs which is applied in computer vision field, and they summarize over 200 papers based on the MRFs. Among them, only seven papers focus on medical images, and no paper is related to pathology image analysis.

Irshad et \textit{al}.~\cite{irshad2013methods} in 2014 summarizes histopathological image analysis methodologies applied in detecting, segmenting and classifying nuclear. According to summarizing table, there are around 100 papers on various image analysis techniques. Among them, there are only two related works using random field models.

Xing et \textit{al}.~\cite{xing2016robust}, in 2016, gives a review concentrating on the newest cell segmentation approaches for microscopy images analysis. In the study, they summarize the CRF models in microscopy image analysis in a separate subsection. This review concludes 326 papers in total, and mainly three papers employ the CRF models for image classification.

Litjens et \textit{al}.~\cite{litjens2017survey} exhibits a review of DL in medical analysis in the year of 2017. This paper provides overviews in a full range of application area, including neuro, retinal, digital pathology and so on. It summarizes over 300 contributions on various imaging modalities. Among them, twelve papers are relate to random field models.

Chang et \textit{al}.~\cite{chang2019artificial} in 2018 presents a survey article based on recent advances in artificial intelligence applied to pathology. In this paper, around 73 papers associated with this topic are concluded. However, only one paper among them mentions the CRFs.

Wu et \textit{al}.~\cite{wu2018image}, in 2019, reviews several modern image labeling methods based on the MRFs and CRFs. In addition, they compare the result of random fields with some classical image labeling methods. However, they give less priority of pathology images. Among 28 papers summarized, only one paper concentrates on medical image analysis. No paper is related to pathology images.

Wang et \textit{al}.~\cite{wang2019pathology} publishes a review in 2019, which explores the pathology image segmentation process using DL algorithm. In this review, the detailed process of whole image segmentation is described from data preparation to post-processing step. In the summary survey, there is only one paper based on random field models.

Yu et \textit{al}.~\cite{yu2019comprehensive}, in 2019, publishes a review that presents the summary of different versions of the CRF models and their applications. This paper classifies application fields of the CRFs into four categories, and discusses their application directions in biomedicine separately. There are 37 papers summarized in that subsection. However, only 20 of them concentrate on medical images, and two of them relate to pathology images.

Li et \textit{al}.~\cite{li2020review}, in 2020 presents a review that summarizes the recent researches related to cervical histopathology image analysis with machine learning (ML) methods. Various machine learning methods are discussed grouped by the application goals in this research. Only two papers using novel multilayer hidden conditional random fields (MHCRFs) are included.

Rahaman et \textit{al}.~\cite{rahaman2020survey}, in 2020, gives a review concentrating on cervical cytopathology image analysis. The algorithms applied in the papers concluded are mainly based on DL methods. However, only one research of the 178 papers that focuses on a local FC-CRF combined with the CNNs is described specifically in application in segmentation part.

In addition to the surveys that have mentioned before, there exists other reviews in related fields, such as the works in~\cite{li1995MRF}, \cite{he2010computer} and~\cite{komura2018machine}.

\begin{figure}[htbp]
  \centering
  \centerline{\includegraphics[width=0.98\linewidth]{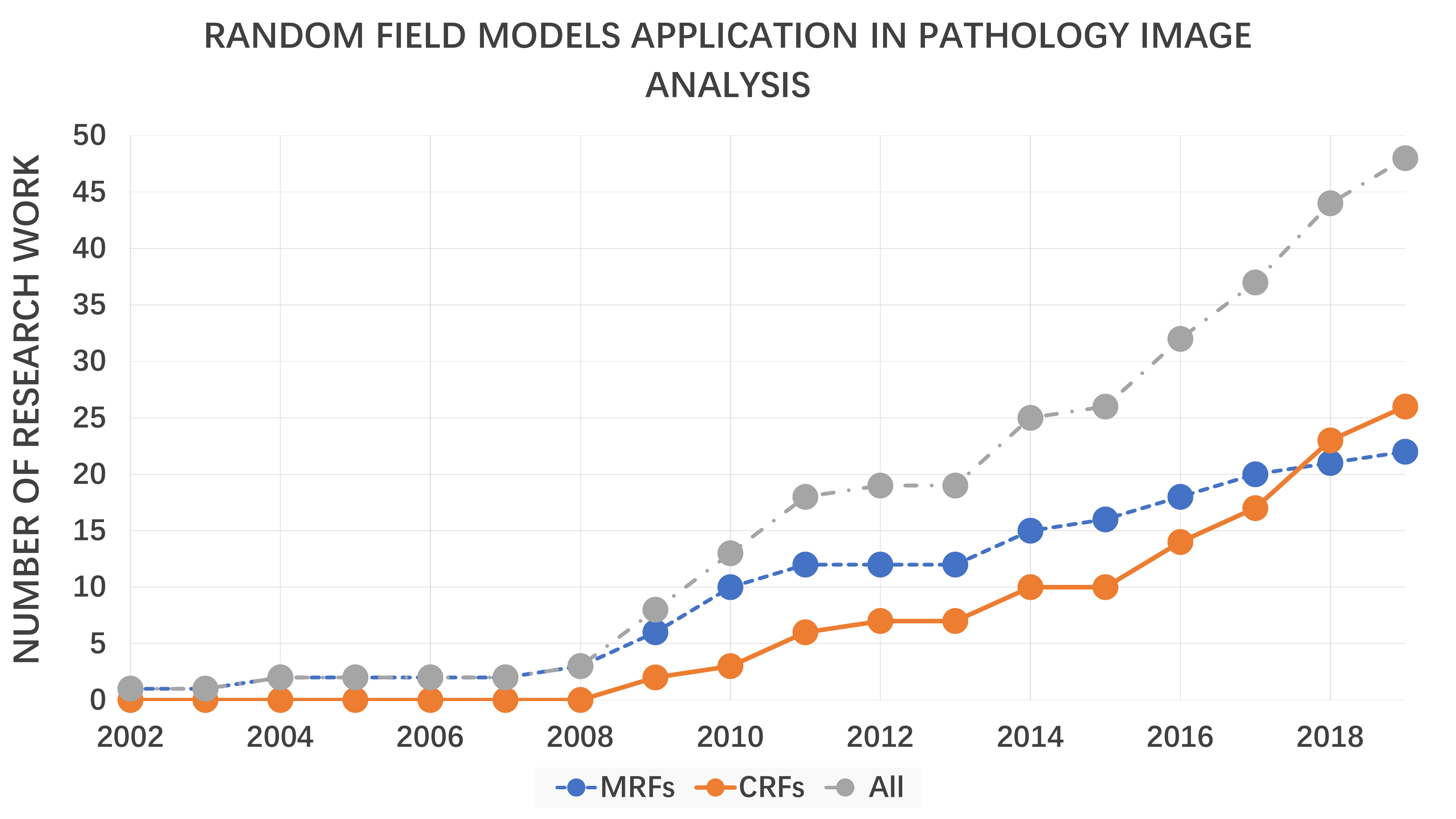}}
\caption{Development trend for MRFs and CRFs methods applied in pathology image analysis.}
\label{fig:ry}
\end{figure}

\begin{figure}[htbp]
\centering
\centerline{\includegraphics[width=0.98\columnwidth]{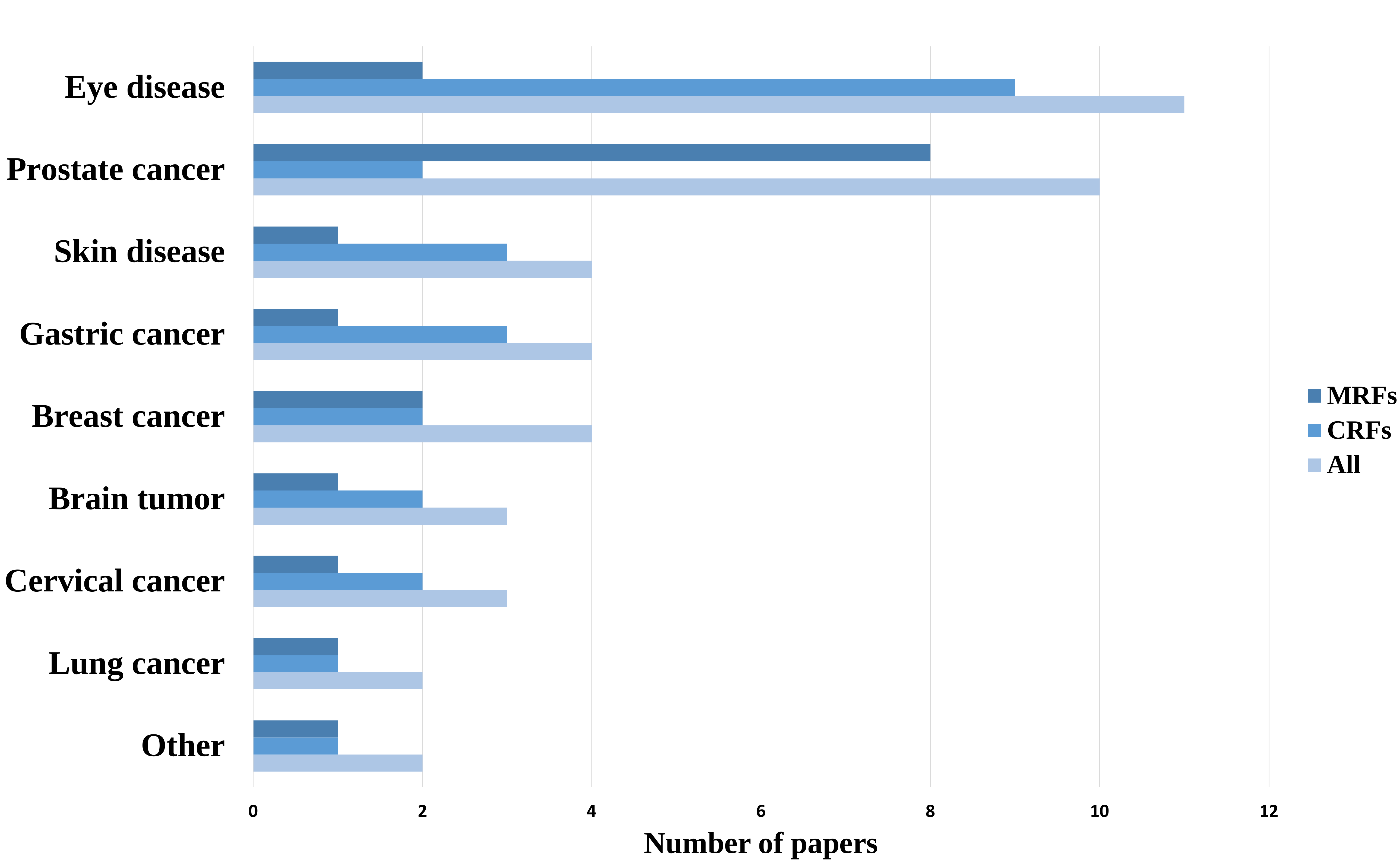}}
\caption{Breakdown of the papers included in this survey in year of medical application area.}
\label{fig:disease-statistics}
\end{figure}

\begin{figure}[htbp]
\centering
\centerline{\includegraphics[width=0.98\columnwidth]{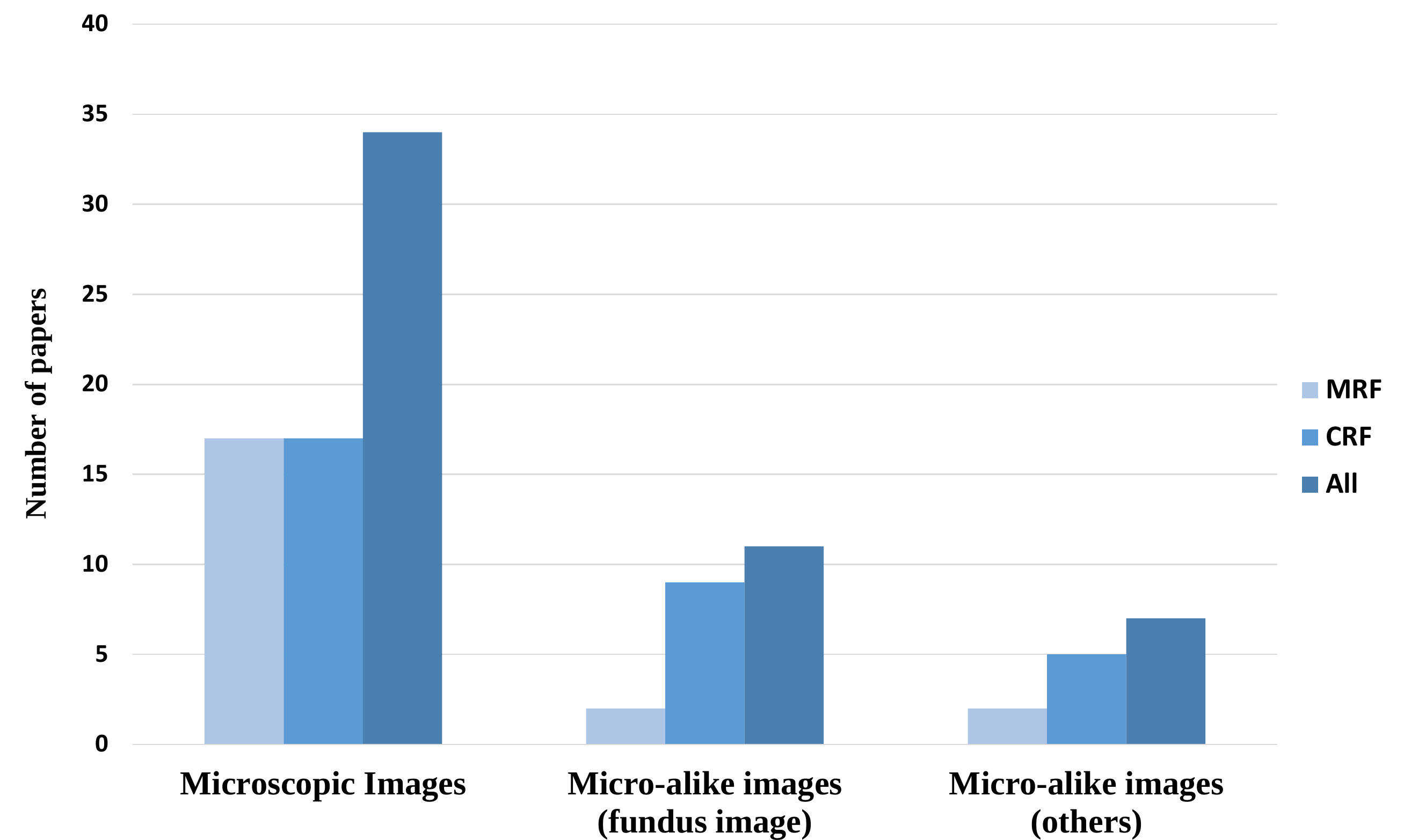}}
\caption{Breakdown of the papers included in this survey in year of medical imaging modality.}
\label{fig:data-kind}
\end{figure}

From the existing survey papers, it is evident that many researchers have paid attention 
to pathology image analysis, or random field models’ application. However, we have not found a single review that concentrates on pathological analysis using the MRF and CRF methods. Therefore, this paper gives a comprehensive review of the relevant work in the past decades. This paper summarizes nearly 40 related works from 2000 to 2019. This is presented in Fig.~\ref{fig:ry}, Fig.~\ref{fig:disease-statistics} 
and Fig.~\ref{fig:data-kind}. Fig.~\ref{fig:ry} illustrates the increasing number of papers that discuss random field models applied in pathology images analysis domain.

From 2002 to 2008, the number of researchers working on the MRFs increased steadily, 
but the figure remained at a low level. The figure of CRFs stayed zero over that period. 
After 2002, they all experienced a rapid upward trend. Compared to the papers focusing 
on the MRFs, that of CRFs grew more rapidly and exceeded its counterpart in 2017. 
Overall, the total number of research papers saw a consistent rise throughout the 
period shown.

\begin{figure*}[!t]
\centerline{\includegraphics[width=1.9\columnwidth]{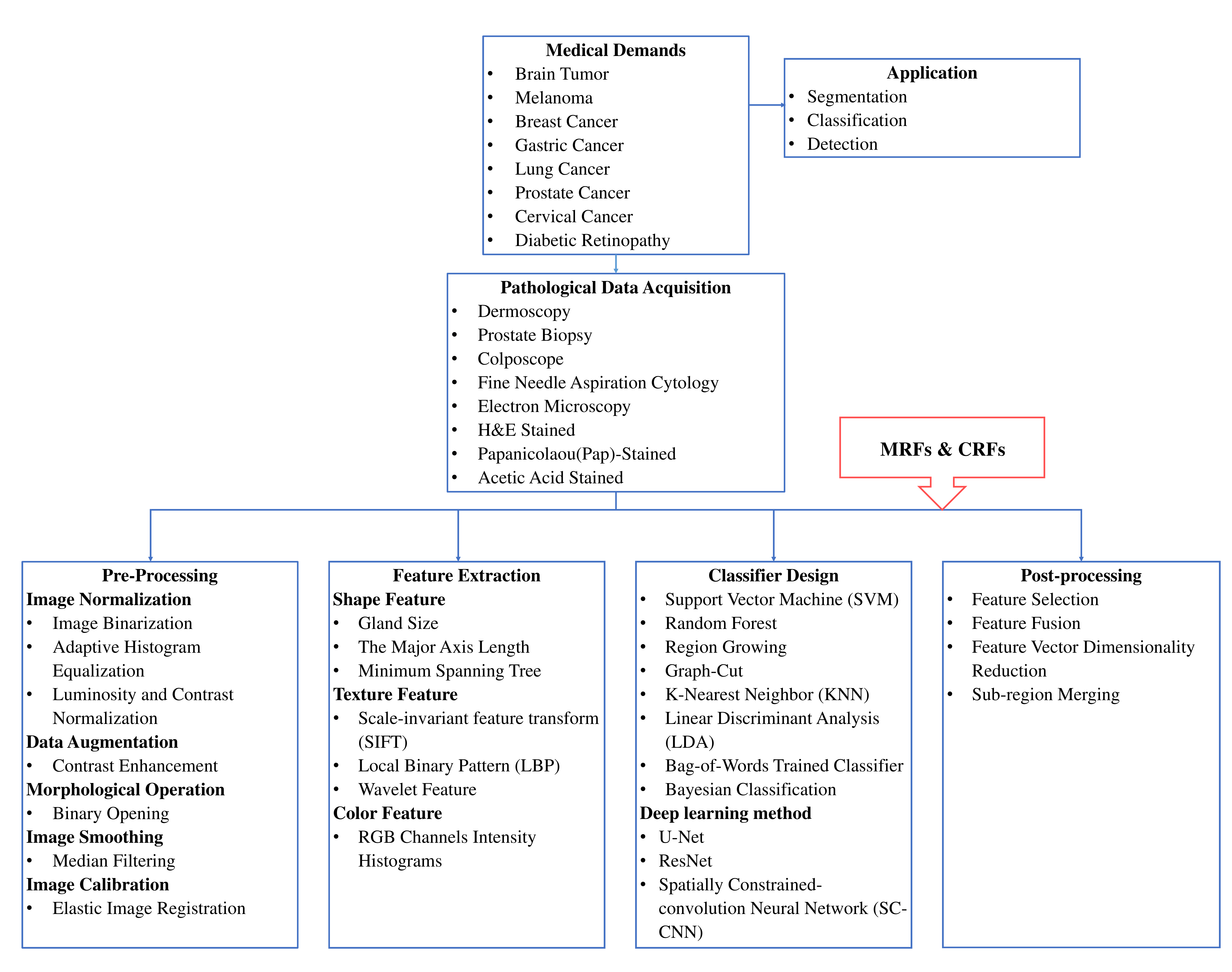}}
\caption{The conclusion of popular machine learning algorithm for pathology image analysis , consisting of image acquisition, image pre-processing, feature representation, classifier design and image post-processing.}
\label{fig:workflow}
\end{figure*}

Concluded from the papers that have been mentioned, a flow chart is given and shown in 
Fig.~\ref{fig:workflow}. It includes the most popular methods in each step that have 
been used in pathology image analysis.

\subsection{Data Description}
\renewcommand\arraystretch{1.9}
\begin{table*}[!t]
    \centering
    \caption{Detailed information of frequently-used retinal datasets.}
    \tiny
    \begin{tabular}{|c|c|c|c|c|}
    \hline
        Datasets & Reference & Download Link & Quantity of dataset & Resolution (pixels) \\ \hline
        DRIVE & \cite{DRIVE}& http://www.isi.uu.nl/Research/Databases/DRIVE/ & 40 & 565 $\times$ 584 \\ \hline
        STARE & \cite{STARE} & http://cecas.clemson.edu/\~ahoover/stare/ & 81 & 700 $\times$ 605 \\ \hline
        CHASEDB1 & \cite{CHASE} & https://blogs.kingston.ac.uk/retinal/chasedb1/ & 28 & 1280 $\times$ 960 \\ \hline
        HRF & \cite{HRF} & https://www5.cs.fau.de/research/data/fundus-images/ & 45 & 3304 $\times$ 2336 \\ \hline
        DRION & \cite{DRION}  & https://zenodo.org/record/1410497\#.X1RFKMgzY2w & 110 & 923 $\times$ 596 \\ \hline
        MESSIDOR & \cite{MESSIDOR} & http://www.adcis.net/en/third-party/messidor/ & 1200 & 1440$\times$ 960, 2240 $\times$ 1488 or 2304 $\times$ 1536  \\ \hline
    \end{tabular}
    \label{Retinaldata}
\end{table*}

It is observed from the reviewed papers that publicly available retinal datasets are 
frequently used to prove the effectiveness of the proposed method or make comparisons 
between different approaches. These datasets are detailed in Table ~\ref{Retinaldata}.

\section{Basic Knowledge of MRFs and CRFs}
\begin{figure*}[!t]
\centering
\centerline{\includegraphics[width=1.9\columnwidth]{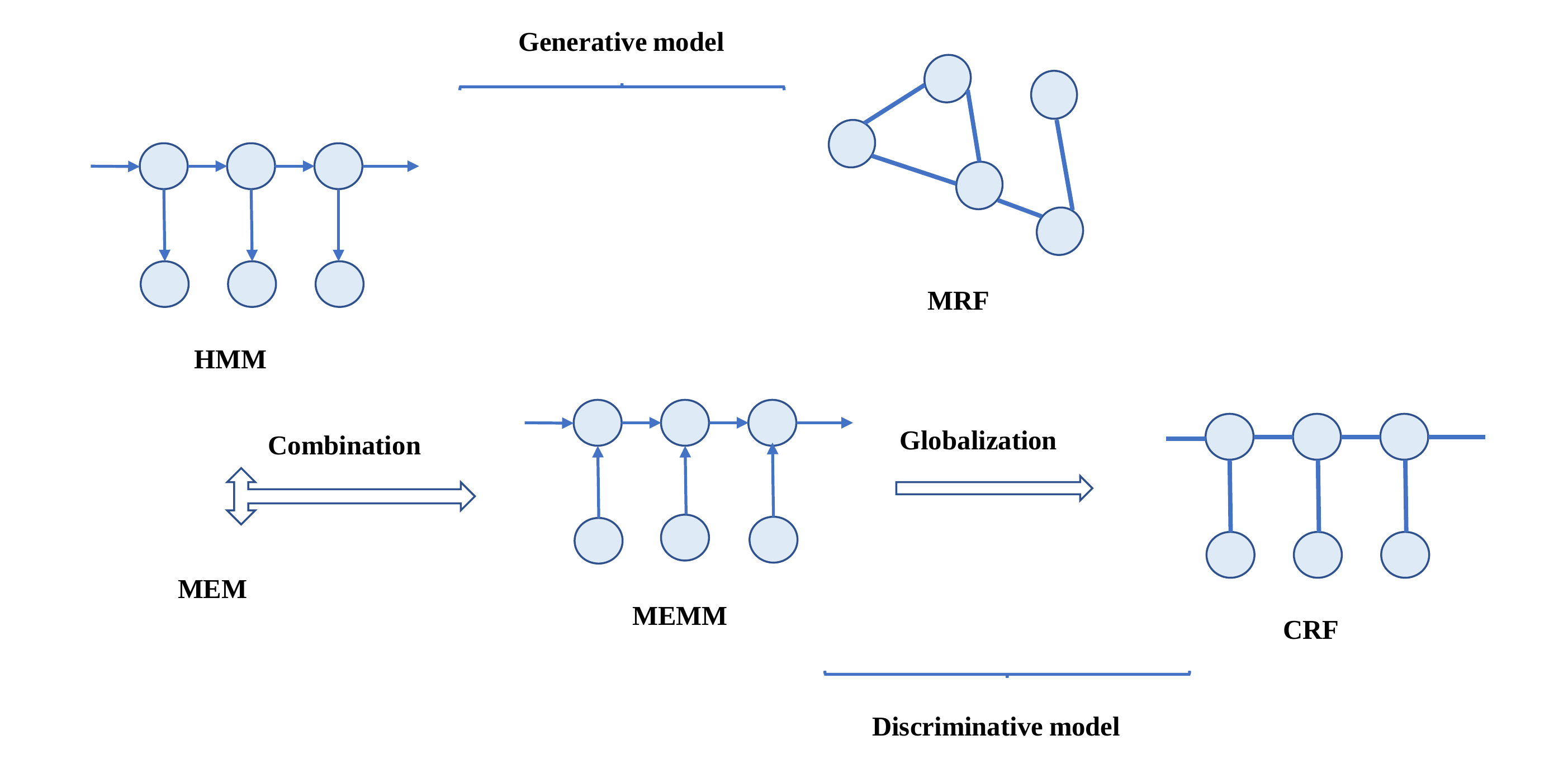}}
\caption{The emergence background of the MRF and CRF models (Maximum Entropy Model (MEM), Maximum Entropy Markov Models (MEMMs), Hidden Markov Model (HMM)).}
\label{fig:ppm}
\end{figure*}

The goal of this section is to provide a formal introduction of the basic knowledge 
of the MRFs and CRFs, including their modeling process, property and inference. 
The background of the two models is shown in Fig.~\ref{fig:ppm}.

\subsection{Basic Knowledge of MRFs}
\subsubsection{Modeling of MRFs}
The random field models' key elements contain the node and the edge in the probabilistic graphical model. The key mathematical symbols during modeling process are listed in Table~\ref{math}. The triplet $(G, \Omega, P)$ is finally defined as the random field model.
\renewcommand\arraystretch{1.5}
\begin{table}[h]
    \centering
    \caption{The definition and brief introduction of the key mathematical symbol in random fields.}
    \scriptsize
    \begin{tabular}{|p{3.5cm}<{\raggedright}|p{4cm}<{\raggedright}|}
    \hline
       \textbf{Mathematical symbol}  & \textbf{Description} \\ \hline
        $S=\{1,2,...,N\}$ & The set containing $N$ sites to be classified \\ \hline
        $s\in S$ & The site belongs to set S \\ \hline
        $X_{s}\in \Lambda\equiv \{ \omega_{1},\omega_{2} ,...,\omega_{L} \}$ & State (class) of each site \\ \hline
        $Y_{s}\in\mathbb{R}^{D}$ & $D$-dimensional feature vector of each site \\ \hline
        $x_{s}\in\Lambda$ & Particular instances of $X_{s}$ \\ \hline
        $y_{s}\in\mathbb{R}^{D}$ & Particular instances of $Y_{s}$ \\ \hline
        $\mathbf{X}=(X_{1},X_{2},...,X_{N})$ & All random variables $X_{s}$  \\ \hline
        $\mathbf{Y}=(Y_{1},Y_{2},...,Y_{N})$ & All random variables $Y_{s}$  \\ \hline
        $\Omega =\Lambda ^{N}$ & The state spaces of $\mathbf{X}$ \\ \hline
        $\mathbb{R}^{D\times N}$ & The state spaces of $\mathbf{Y}$ \\ \hline
        $\mathbf{x}=(x_{1},x_{2},...,x_{N})\in\Omega$ & Instances of $\mathbf{X}$ \\ \hline
        $\mathbf{y}=(y_{1},y_{2},...,y_{N})\in\mathbb{R}^{D\times N}$ & Instances of $\mathbf{Y}$ \\ \hline
        $G=\{S,E\}$ & An undirected graph structure on the sites, $S$ and $E$ are the vertices (sites) and edges \\ \hline
        $\eta_{s}$ & A neighborhood set contains all sites that share an edge with $s$ \\ \hline
        $P$ & A probability measure defined over $\Omega$ \\ \hline
        $(G, \Omega, P)$ & Random field \\ \hline
    \end{tabular}
    \label{math}
\end{table}

\subsubsection{Property of MRFs}
If the local conditional probability density 
functions (LCPDFs) of the random field $(G, \Omega, P)$ conform to the characteristics of the Markov property shown in Eq.~\ref{equ:1}, it will be defined as an MRF.
\begin{equation}
P(x_{s}\mid \mathbf{x}_{-s})=P(x_{s}\mid \mathbf{x}_{\eta_{s}}),
\label{equ:1}
\end{equation}
where 
$\mathbf{x}_{-s}=(x_{1},...,x_{s-1},x_{S+1},...,x_{N})$,\\
$\mathbf{x}_{\eta_{s}}=(x_{\eta_{s}(1)},x_{\eta_{s}(\left | \eta_{s}\right |)})$, and $\eta_{i}\in S$ is the $i^{th}$ element of the set $\eta_{s}$.
Thus, the forms of the LCPDFs are simplified by Markov property~\cite{monaco2009probabilistic}.

\subsubsection{Inference of MRFs}
Given an observation of the feature vectors $Y$, the states $X$ is to be estimated. 
The preferred method is an MAP estimation which entails maximizing the following 
quantity defined in Eq.~\ref{equ:2} over all $x\in\Omega$: 
\begin{equation}
\begin{split}
P(x\mid y)=\frac{P(y\mid x)P(x)}{P(y)}\\
\propto P(y\mid x)P(x)
\label{equ:2}
\end{split}
\end{equation}

The first term of Eq.~\ref{equ:2} indicates the impact of the feature vectors. Given their association $X_{s}$, to simplify the problem, it is possible to assume that all $Y_{s}$ are conditionally independent and that they have the same distribution. 
This means that if the class $X_{s}$ of site $s$ is known, then the classes and characteristics of the remaining sites do not offer additional information when evaluating $Y_{s}$, and the conditional distribution of $Y_{S}$ is the same for all $s\in S$. Thus, Eq.~\ref{equ:3} 
is derived and shown as follows: 
\begin{equation}
\begin{split}
P(y\mid x)=\prod_{s\in S} P(y_{s}\mid x_{s})\\
=\prod_{s\in S} p_{f}(y_{s}\mid x_{s}),
\label{equ:3}
\end{split}
\end{equation}

The use of the single PDF $p_{f}$ in Eq.~\eqref{equ:3} demonstrates that $P(y_{s}\mid x_{s})$ is uniformly distributed across $S$. The second term in Eq.~\ref{equ:3} indicates the impact of the class labels. Generally, it is intractable to model this high-dimensional PDF. Nevertheless, its formulation will be obviously simplified if the Markov property is assumed.

Hammersley-Clifford (Gibbs-Markov equivalence) theorem shows the connection between the Markov property and the JPDF of $\mathbf{X}$, which indicates that 
a random field $(G, \Omega, P)$ with $P(\mathbf{x})>0$ for all $\mathbf{x}\in\Omega$ satisfies the Markov property when it can be expressed as a Gibbs distribution in Eq.~\ref{equ:4}: 
\begin{equation}
P(\mathbf{x})=\dfrac{1}{Z}\prod_{C}V_{C}(\mathbf{x}),
\label{equ:4}
\end{equation}
where $V_{C}(\mathbf{x})$ denotes the potential function on the clique 
$\textit{C}$. $Z=\sum_{\mathbf{x}\in\Omega}\prod_{C}V_{C}(\mathbf{x})$ is the 
normalization factor. A clique $\textit{C}$ is a subset of the vertices, and $C \subseteq S$, such that every two distinct nodes are adjacent in an undirected graph $\textit{G} = (\textit{S},\textit{E} )$~\cite{xu2017connecting}.

\subsection{Basic Knowledge of CRFs} 
Supposing $\textbf{Y}$ is a random variable of the data sequences to be labeled, 
and $\textbf{X}$ is a random variable representing the corresponding label sequences. If $G=(S, E)$ is a undirected graph such that $\textbf{X}=(\textbf{X}_{s})_{s\in S}$, 
so that $\textit{X}$ is indexed by the vertices of $G$. Then assuming $(\textbf{X}, \textbf{Y})$ 
is a CRF, when condition on $\textbf{Y}$, the random variables $\textbf{X}_{s}$ 
obey the Markov property mentioned before. Compared with the MRF model, a conditional 
model $p(\textbf{X}\mid \textbf{Y})$ from paired observation and label sequences, and 
not explicitly model the marginal $p(\textbf{Y})$~\cite{lafferty2001conditional}. 
The CRF can be represented by Eq.~\ref{equ:5} as follows. 
\begin{equation}
\begin{split}
P(x\mid y)=\frac{1}{Z(y)}exp(\sum_{i,k}^{}\lambda_{k}t_{k}(x_{i-1},x_{i},y,i)\\
+\sum_{i,j}^{}\mu _{j}s_{j}(x_{i},y,i))\\
Z(y)=\sum_{x}^{}exp(\sum_{i,k}^{}\lambda_{k}t_{k}(x_{i-1},x_{i},y,i)\\
+\sum_{i,j}^{}\mu _{j}s_{j}(x_{i},y,i))
\label{equ:5}
\end{split}
\end{equation}

In this form, $t_{k}$ and $s_{j}$ are the feature function depending on the positions, 
where $t_{k}$ is the transition feature function defined on the edge. It represents a feature in transmission from one node to the next and is dependent on the current positions. $s_{j}$ denotes the state feature function that is dependent on the current location, representing the characteristics of the node. $\lambda_{k}$ and $\mu _{j}$ are learning parameters that is going to be estimated, and $Z(y)$ denotes the normalization factor, which performs a sum over all possible output sequences~\cite{yu2019comprehensive}.

\subsection{Optimization Algorithm} 
In this subsection, some representative optimization algorithms frequently used 
in the related works are introduced. Most of them are iterative models applied for the observation estimation from known distributions.

\subsubsection{Expectation-maximization (EM)}
Mixed models (such as random field models) can be fitted with maximum likelihood through the EM method without data. The algorithm first takes the initial model parameters as a priori and then uses them to estimate the missing data. Once achieving the complete data, the likelihood expectation maximization will be repeatedly performed to estimate the parameters of the model. The algorithm contains the expectation and maximization step. The optimal parameters estimation is inferred in each step, thereby achieving the best data~\cite{wu2008top}.

\subsubsection{Iterative Conditional Modes (ICM)}
ICM is an iterative and intuitive approach that utilizes knowledge of neighborhood system for  optimal solution inference. $P(X\mid Y)$ or a random selection serves as initial conditions of the method. If the new solution owns the lowest energy, ICM attempts to use the current solution to upgrade the label at each position. When the energy at any position cannot be further reduced, the algorithm converges~\cite{mungle2017mrf}.

\subsubsection{Simulated Annealing}
Simulated Annealing is a classical technique for optimization. It simulates the 
process of annealing to find global minimums or maximums from local minimum or 
maximums~\cite{mungle2017mrf}.

\section{MRFs for Pathology Image Analysis}
\label{sec:MRFs}
In this section, the related research papers focusing on pathology image analysis using the MRF methods are grouped into two basic categories and surveyed, including application in segmentation and other tasks. Finally, a conclusion of method analysis is presented in the last paragraph.
\subsection{Image Segmentation Using MRFs}
In the research of tuberculosis disease, the quantification of immune cell recruitment is necessary. Considering that fact, in~\cite{meas2002color}, an automatic cell counting method for histological image analysis containing color image segmentation is introduced. In this research, a new clustering approach based on a simplified MRF model is developed, which is called the MRF clustering (MRFC) method. It uses the Potts model as a basic model, which is defined in eight connexity using second order cliques and is able to handle both color and spatial information. They also complement the MRFC with a watershed on the binary segmentation result of aggregated zones (whose size is higher than a threshold value). This method uses seven groups of mouse lung slice images for testing and gets a cell counting accuracy of 100\% in a group containing 23 images.

\begin{figure}[htbp]
  \centering
  \centerline{\includegraphics[width=0.98\linewidth]{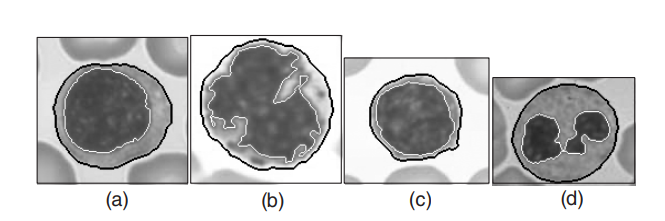}}
\caption{Segmentation results without tangent red blood cell. This figure corresponds to Fig.4 in \cite{won2004segmenting}.}
\label{fig:segmenting}
\end{figure}

An essential task for hematologists and pathologists is to isolate the nucleus and cytoplasmic areas in images of blood cells. This paper introduces a three-step image segmentation method in~\cite{won2004segmenting}. In the first step, an initial segmentation is completed using a Histogram Thresholding method. In the second step, a Deterministic Relaxation method, where  the MRFs is utilized to formulate the MAP criterion, is adopted to smooth the noise area in the last step. Finally, a separation algorithm consisting of four steps is used: boundary smoothing, concavities detection, searching a pair of contour to be connected, and  deleting the false subregion. The proposed segmentation algorithm is employed to 22 cell images, containing four different kinds, and it yields 100\% correct results (the illustration of the experimental results is presented in Fig.~\ref{fig:segmenting}).

In~\cite{zou2009Fuzzy}, a couple method of the MRF and fuzzy clustering is adopted to express the adapt function and segment pathological images. In addition, particle swarm optimization (PSO) is added to the fuzzy clustering method. Thus, this model has the strong capability of noise immunity, quick convergence rate and powerful ability of global search. On a 821-cell dataset, an accuracy value of 86.45\% is finally achieved.

\begin{figure}[htbp]
  \centering
  \centerline{\includegraphics[width=1.1\linewidth]{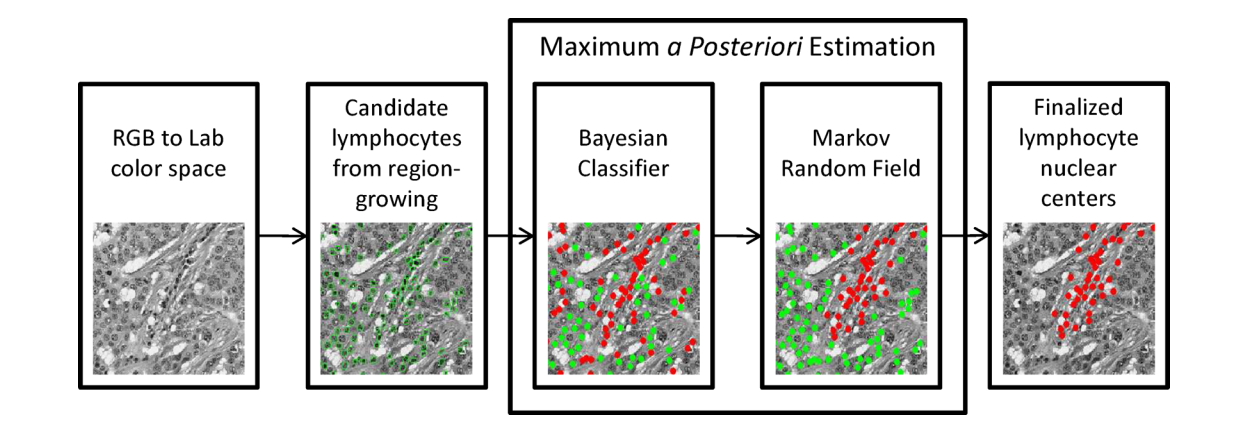}}
\caption{The flowchart explains the main five steps in the automated lymphocyte detection framework. This figure corresponds to Fig.3 in~\cite{basavanhally2009computerized}. }
\label{fig:MRF2009}
\end{figure}

In order to detect and grade the degree of lymphocyte infiltration (LI) in HER2+ breast cancer histopathology images automatically, a quantitative CAD system is developed in~\cite{basavanhally2009computerized}. The lymphocytes are located with the integration of region growing and the MRF method first (the flow chart is shown in Fig.~\ref{fig:MRF2009}). The Voronoi diagram, Delaunay triangulation diagram, and minimum spanning tree diagram are applied. The center of lymphocytes detected separately serve as the vertex. After feature extraction from each sample, the non-linear dimensionality reduction scheme is applied to produce dimensionality reduction embedded space feature vector. Finally, an SVM classifier is adopted to distinguish the samples with high or low LI  in the dimensionality reduction embedded space. In the first step, after Bayesian Modelling of LI via MAP estimation, an MRF model defines the prior distribution $p(x)$, using ICM as an optimization algorithm to assign a hard label to each random variable. Afterward, each region is classified as either breast cancer or lymphocyte nucleus. In this experiment, totally 41 H\&E stained breast biopsy samples are tested, yielding an accuracy value of 0.9041.

\begin{figure}[htbp]
  \centering
  \centerline{\includegraphics[width=0.98\linewidth]{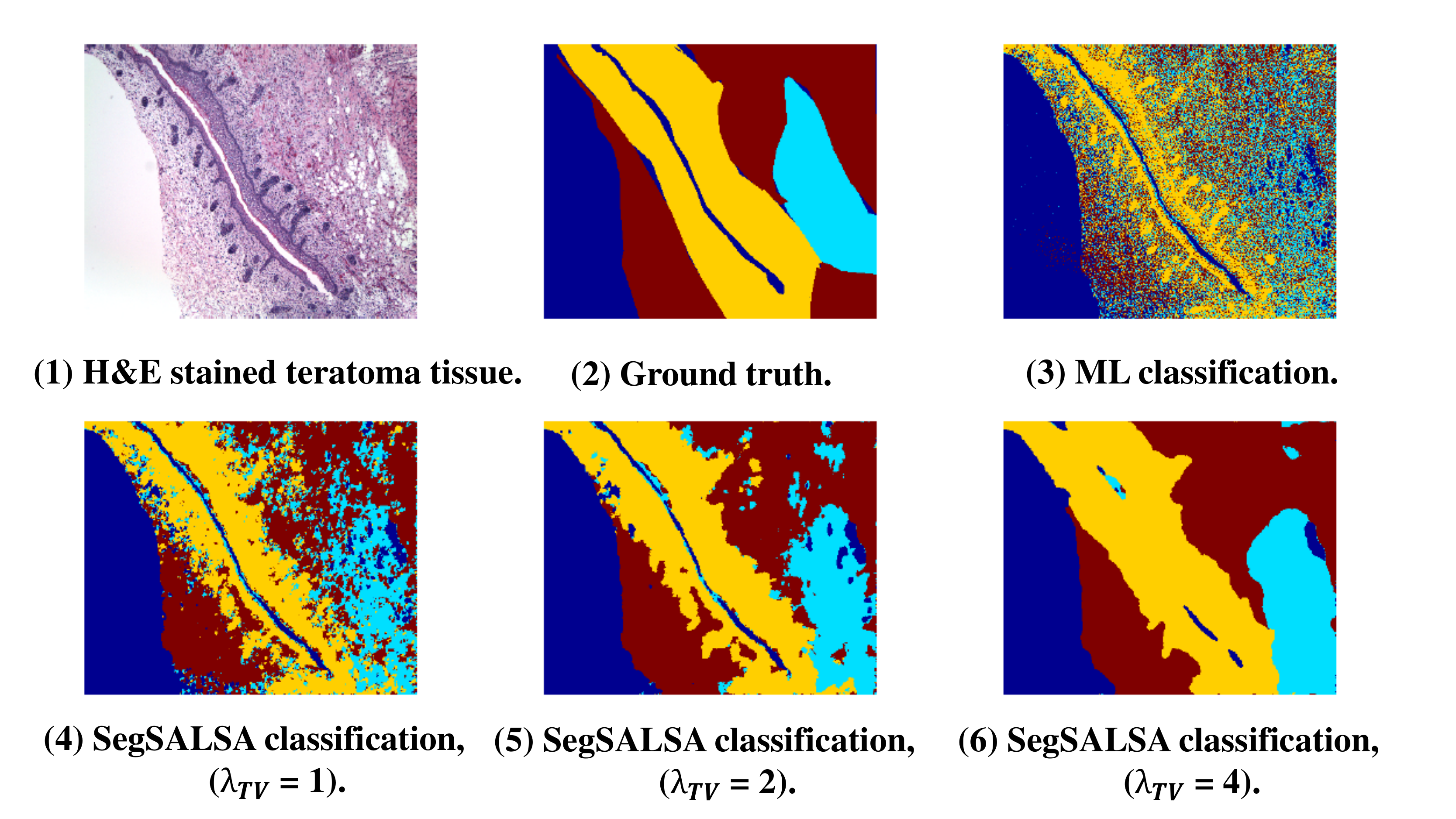}}
\caption{H\&E stained teratoma tissue sample consisting four classes of cell , which is imaged under 40X magnification microscope. The detailed information of this figure can be seen in Fig.4 in \cite{bioucas2014alternating}. }
\label{fig:SAL}
\end{figure}

In~\cite{bioucas2014alternating}, a four-step image segmentation process is employed to classify four categories of teratoma tissues. First, the image segmentation process is formulated in the Bayesian framework. Second, a set of hidden real-valued random fields designing a given segmentation probability is introduced. In order to produce smooth fields, a Gaussian MRF (GMRF) prior is assigned to reformulate the original segmentation problem. This method's salient feature is that the original discrete optimization is transformed into a convex optimization problem, which makes it easier to use related tools to solve the problem accurately. Third, the form of total variation of isotropic vectors is adopted. Finally, aiming to conquer the convex optimization problem that makes up the MAP inference of hidden field, the Segmentation via a Constrained Split Augmented Lagrangian Shrinkage Algorithm (SegSALSA) segmentation is introduced. As Fig.~\ref{fig:SAL} shown, the proposed system with SegSALSA finally yields an accuracy value of 0.84.

\begin{figure}[htbp]
  \centering
  \centerline{\includegraphics[width=0.98\linewidth]{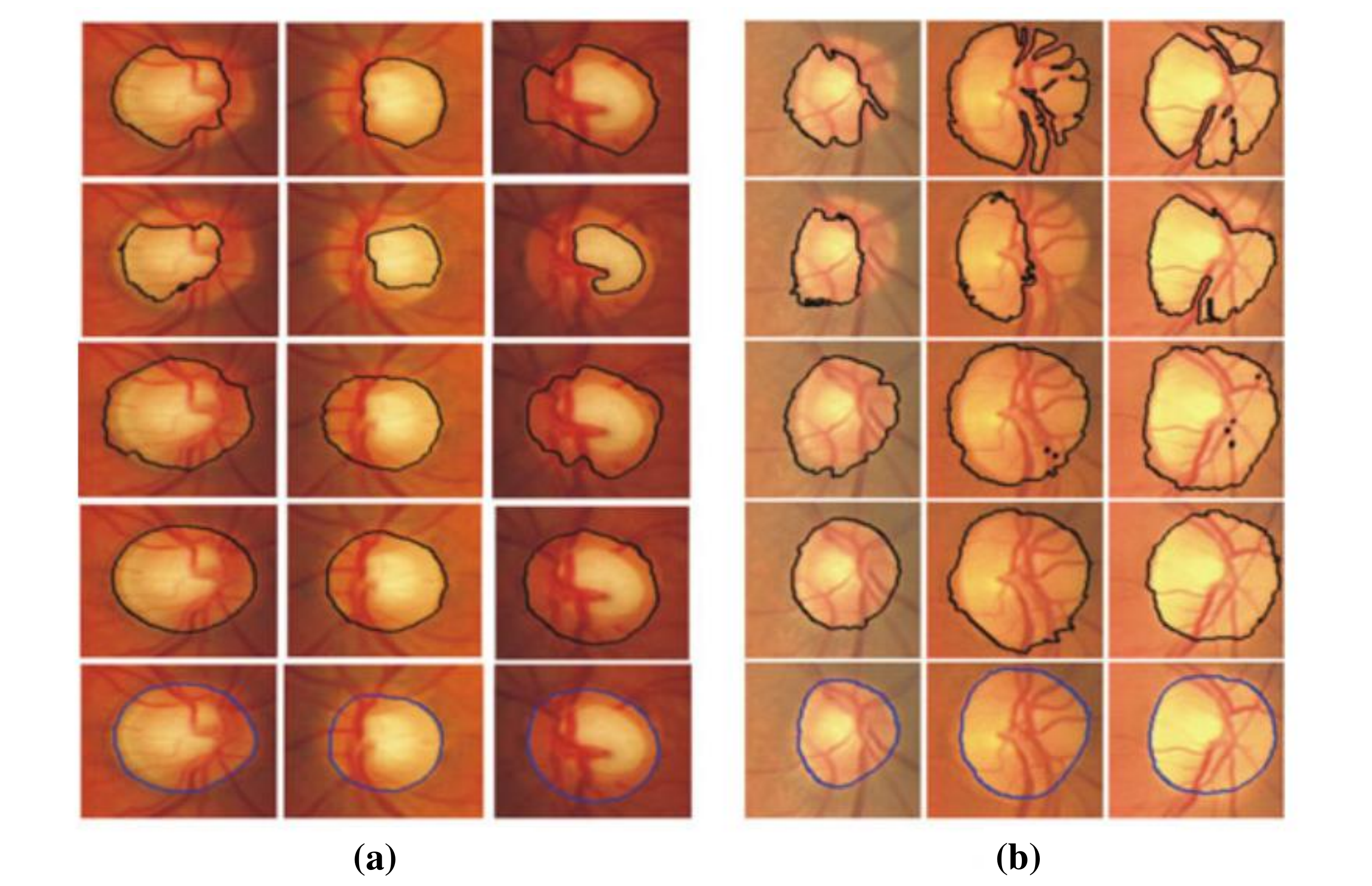}}
\caption{The illustration of optic disk segmentation results tested on DIARETDB1 and DRIVE dataset. This figure corresponds to Fig.14 in \cite{SalazarSegmentation}. }
\label{fig:disk}
\end{figure}

The morphology of retinal vessels and optic disc is an important consideration to the diagnosis of many retinal diseases, such as diabetic retinopathy (DR), hypertension, glaucoma. Given that fact, in~\cite{SalazarSegmentation}, an MRF image reconstruction method is applied to segment the optic disk. The first step is to extract the retina vascular tree applying the graph cut technique so that the location of the optic disk can be estimated on the basis of the blood vessel. Afterward, the MRF method is adopted to define the location of the optic disk. It plays a significant role in eliminating the vessel from the optic disk area and meanwhile avoids other structural modifications of the image. Fig.~\ref{fig:disk} (a) and (b) illustrate the optic disk segmentation results, compared with other prevalent methods. On the DRIVE dataset, an average overlapping ratio 0.8240, mean absolute distance 3.39, and sensitivity 0.9819 are finally achieved.

In \cite{liu2015study}, aiming to improve Melanoma’s early diagnosis accuracy, an overall process research based on dermoscopy images is proposed, including image noise removal, lesion region segmentation, feature extraction, recognition of skin lesions and its classification. Lesion region segmentation is the first and essential step, where the image noise is removed first using contrast enhancement method, threshold and morphological method. In the main part, a fusion segmentation algorithm applying the MRF segmentation framework is introduced to develop the robustness of a single segmentation algorithm. The fusion strategy transforms the optimal fusion segmentation problem into the problem of minimizing the multi-dimensional space energy composed by the results of four segmentation algorithms (Statistical Region Merging, Adaptive Thresholding, Gradient Vector Flow Snake and Level Set). 1039 RGB images derived from two European universities are used for training and the overall process research finally achieves classification accuracy 94.49\%, sensitivity 95.67\% and specificity 94.31\%.

\begin{figure}[htbp]
  \centering
  \centerline{\includegraphics[width=0.98\linewidth]{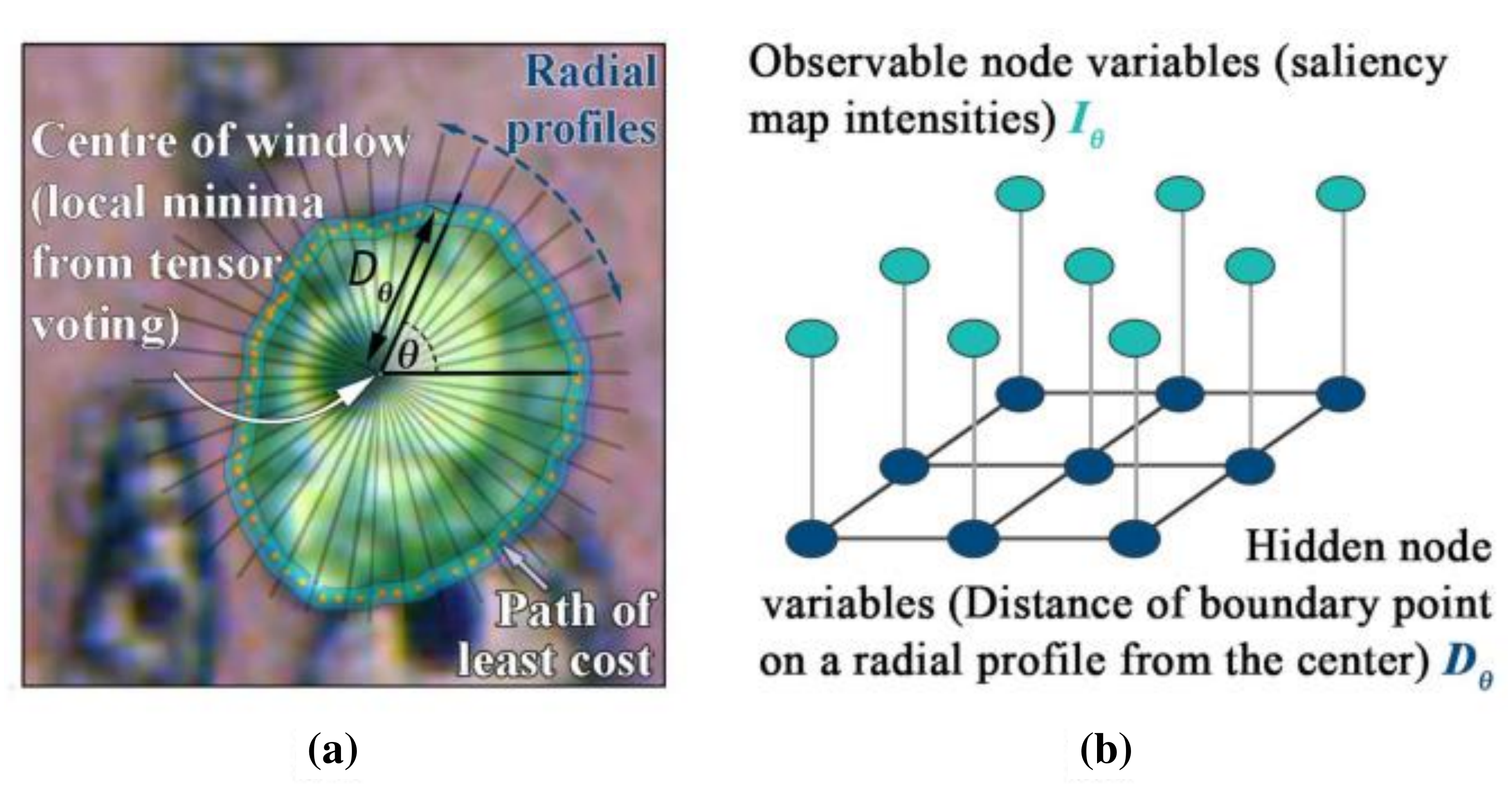}}
\caption{a) Graphical illustration of the essential step in boundary delineation problem. b) Probabilistic 
graphical model of the MRF applied in this paper. The detailed information of this figure can be seen in Fig.5 in ~\cite{paramanandam2016automated}. }
\label{fig:PCA}
\end{figure}

Nuclei segmentation is one of the essential steps for breast histopathology image analysis. To detect the nuclei boundary, a four-stage procedure is proposed in~\cite{paramanandam2016automated}. In the preprocessing step, the enhanced grayscale images are obtained by using the principal component analysis (PCA) method. Second, the nuclei saliency map is constructed applying tensor voting. Thirdly, loopy belief propagation on the MRF model is applied to extract the nuclei boundary. At this stage, researchers determine the nuclear boundary by a set of radial profiles with equal arc-length intervals from the window edge to the window center, which is illustrated in Fig.~\ref{fig:PCA}(a)). The MRF observable node variable is the intensity value in the polar form of the nuclear significance graph. Meanwhile, the hidden node variable is the radial profile of the nuclear boundary points, which is shown in Fig.~\ref{fig:PCA}(b). Finally, spurious nuclei are detected and removed after threshold processing. In a breast histopathology image containing 512 nuclei, the proposed system gets a nucleus segmentation precision of 0.9657, recall of 0.7480, and Dice coefficient of 0.8830, respectively.

In~\cite{Razieh2016Microaneurysms}, an MRF-based novel microaneurysms (MA) segmentation method is developed for DR detection in the retina, where the vessel network is first removed using a contrast enhancement method, then the MA candidates are extracted using local applying of the MRF. Lastly, an SVM classifier is designed to identify true MAs employing 23 features. Based on the motivation of our review, we only concentrate on the MA candidate extraction technique. The EM algorithm is applied for the mean and variance estimation of each category. Moreover, the main equation optimization is conducted with a simulated annealing algorithm. But the MRFs have two limitations: Firstly, the number of regions to be selected after segmentation by the MRFs is so large that the false positive rate may increase; second, the size of some MAs might change, so they cannot represent true MAs anymore. To overcome these limitations, the region growing algorithm is used, where some regions are finally marked as non-MAs and precluded from becoming candidate regions. The authors manage to achieve an average sensitivity value of 0.82 on the publicly available dataset DIAREDB1.

In \cite{zhao2016automatic}, a novel superpixel-based MRF framework is proposed for color cervical smear image segmentation, where the SLIC algorithm generates the superpixels and a feature containing a 13-dimensional vector is extracted from each superpixel first. Second, initial segmentation results are provided by $k$-means++.  Finally, images are modeled as an MRF, making the edges smoother and more coherent to semantic objects. An iterative adaptive classified algorithm is adopted for parameter estimation. Moreover, This study introduces a gap search algorithm to speed up iteration, which only updates the energy of critical local areas, such as edge gaps. The best results are achieved on Herlev public dataset, yielding 0.93 Zijdenbos similarity index (ZSI) \cite{gencctav2012unsupervised} of the nuclei segmentation.

In~\cite{mungle2017mrf}, an MRF-ANN algorithm is developed for estrogen receptor scoring using breast cancer immunohistochemical images, where the white balancing method is first applied for color normalization, then the MRF model with EM optimization is adopted to segment the ER cells. In addition, $k$-means clustering is utilized to obtain the initial labels of the MRF model. Lastly, based on the pixel color intensity feature, an artificial neural network (ANN) is employed to obtain an intensity-based score of ER cells. 
Using the standard Allred scoring system, the final ER score is calculated by adding intensity and proportion scores (calculating the percentage of ER-positive cells from the cell count). On a dataset from a medical center in India, the proposed segmentation method finally achieves an F-measure of 0.95 in tissue sections of 65 patients.

In~\cite{dholey2018combining}, a CAD framework is introduced for cancerous cell nuclei analysis. The images are obtained from pap-stained microscopic images of lung Fine Needle Aspiration Cytology (FNAC) sample, which is essential for lung cancer diagnosis. First, edge-preserving bilateral filtering is used for noise removal. Afterward, the Gaussian mixture model-based hidden Markov random field model is applied to segment the nucleus. Then, the bag-of-visual words model is applied to classify the kernel. The scale-invariant feature transformation features are extracted from the candidate kernel, and the random forest classifier model is trained with these features. A hidden Markov random field, as well as its Expectation-Maximization, is employed in the segmentation step to finding out the unknown parameters in the potential function. This algorithm needs morphological post-processing, including morphological opening operation, watershed algorithm, and connected components labeling method. The segmentation process yields a specificity and sensitivity value of 97.93\% and 98.88\%.

A two-level segmentation algorithm based on spatial clustering and the HMRFs is proposed in~\cite{su2019cell} to improve the segmentation accuracy of cell aggregation and adhesion region. First, $k$-means++ clustering is employed to obtain the initial labels of MRFs based on the color feature of pixels in the Lab color space. Second, the spatial expression model of the cell image is constructed by the HMRF, which considers the spatial constraint relation in order to limit the influence of isolated points and smooth segmentation areas at the same time. Finally, the EM algorithm is adopted for model parameter optimization. Meanwhile, the results are finally refined by the iterative algorithm. The experiment is based on the 61 bone marrow cell images from Moffitt Cancer Center, and after ten iterations, the proposed method yields an accuracy value of 0.9685.

\subsection{Other Applications Using MRFs}
\subsubsection{Prostate Cancer Detection from a US Research Team}

A joint research group from the USA, leading by researchers from Rutgers University and the University of Pennsylvania, develops a serial work about Computer-Aided Detection of Prostate Cancer (CaP) on Whole-Mount Histology. These researches share almost the same procedure, and the MRFs are mostly used in the classification part, which shows exciting performance improvement.

In~\cite{monaco2008detection}, a CAD algorithm is developed to detect the CaP in low resolution whole-mount histological sections (WMHSs). In addition to the area of the glands, another distinguishing feature of glandular cancers is that they are close to other cancerous glands. Therefore, the information in these glands is modeled using the MRF model. The process of the CAD algorithm can be concluded into three steps: First, the region growing algorithm is applied to gland segmentation in the luminance channel of H\&E stained samples. Second, the system calculates the morphological characteristics of each gland and then classifies them through Bayesian classification to mark the glands as malignant or benign. Third, the labels obtained in the last process are applied in MRF iteration to generate the final label. In this step, unlike most of the MRF strategies (such as the Potts model), which rely on heuristic formulations, a novel methodology allowing the MRFs to be modeled directly from training data is introduced. The proposed system is tested in four H\&E stained prostate WMHSs, achieving a sensitivity and specificity value of 0.8670 and 0.9524, respectively, for cancerous area detection.

 As an extension of this work, in~\cite{monaco2009weighted}, aiming to solve the disadvantages of the traditional random fields that most of them produce a single, hard classification at a static operating point, the weighted maximum a posteriori (WMAP) estimation and weighted iterated conditional modes (WICM) are introduced. The use of these two algorithms proves to have good performance in 20 WMHSs from 19 patients’ images.
 
\begin{figure}[htbp]
  \centering
  \centerline{\includegraphics[width=0.98\linewidth]{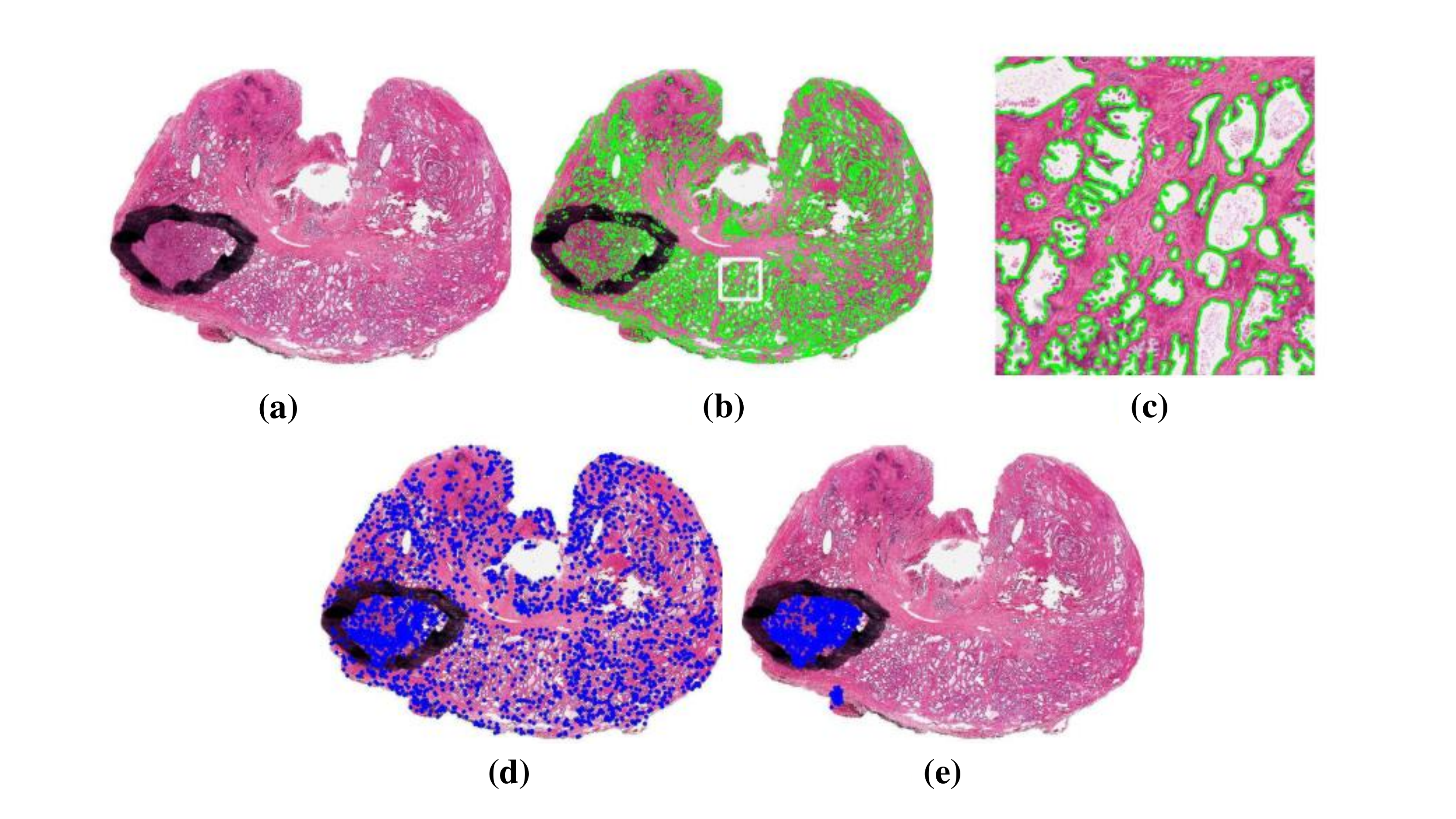}}
\caption{The illustration of ground truth and segmentation performance of the mentioned methods. The detailed information of this figure can be seen in Fig.3 in~\cite{monaco2009probabilistic}.} 
\label{fig:monaco}
\end{figure}

Based on the work above, in~\cite{monaco2009probabilistic} and~\cite{monaco2010high}, the probabilistic pairwise Markov models (PPMMs) are presented. Compared to the typical MRF models, PPMMs, using probability distributions instead of potential functions, have both more intuitive and expansive modeling capabilities. In Fig.~\ref{fig:monaco}, an example of the WMHS image detection steps is shown. In the experiment in \cite{monaco2009probabilistic}, 20 prostate samples obtained from 19 patients in two universities are used for testing. When specificity is held fixed at 0.82, the sensitivity and area of the ROC curve (AUC) of the PPMMs increase to 0.77 and 0.87, respectively (compared to 0.71 and 0.83 using Potts model). 

In~\cite{monaco2010high}, as a supplement to the previous research, the research team hopes to develop a comprehensive and hierarchical algorithm that can finish cancerous areas detection at high speed under a lower resolution, and then refine the results at a higher resolution, and finally perform Gleason classification. The researchers use this method to detect 40 H\& E stained tissue sections of 20 patients undergoing radical prostatectomy in the same hospital, and find that the sensitivity and specificity of CaP detection are 0.87 and 0.90, respectively.

It is found that most of the published researches limit the MRF performance to a single, static operating point. To solve this limitation, in~\cite{monaco2011weighted}, weighted maximum posterior marginals (WMPM) estimation is developed, whose cost function allows the penalty for being classified incorrectly with certain classes to be greater than for other classes. The realization of WMPM estimation needs to calculate the posterior marginal distribution, and the most popular solution is the Markov Chain Monte Carlo (MCMC) algorithm. As an extension of the MCMC method, EMCMC has the same simple structure but better results. It is applied for accurat posterior marginals estimation. This data set includes 27 digitized H\&E stained tissue samples from 10 patients with RPs. By quantitatively comparing the ROC curves of the E-MCMC method and other MCMC methods, experiments prove that this method has better performance.

Based on the work above, a system for detecting regions of CaP using the color fractal dimension (CFD) is established in~\cite{yu2011detection}. In order to find the most suitable hyper-rectangular boundary size for each channel, the researchers improve the traditional CFD algorithm and analyze the values of the three channels in the image separately. And then, the authors combine the probability map constructed via CFD with the PPMM introduced in the above research. In the experiment, an AUC of 0.831 is achieved on a dataset from 10 patients, containing 27 H\&E stained histological sections.

\begin{figure}
  \centering
  \centerline{\includegraphics[width=0.98\linewidth]{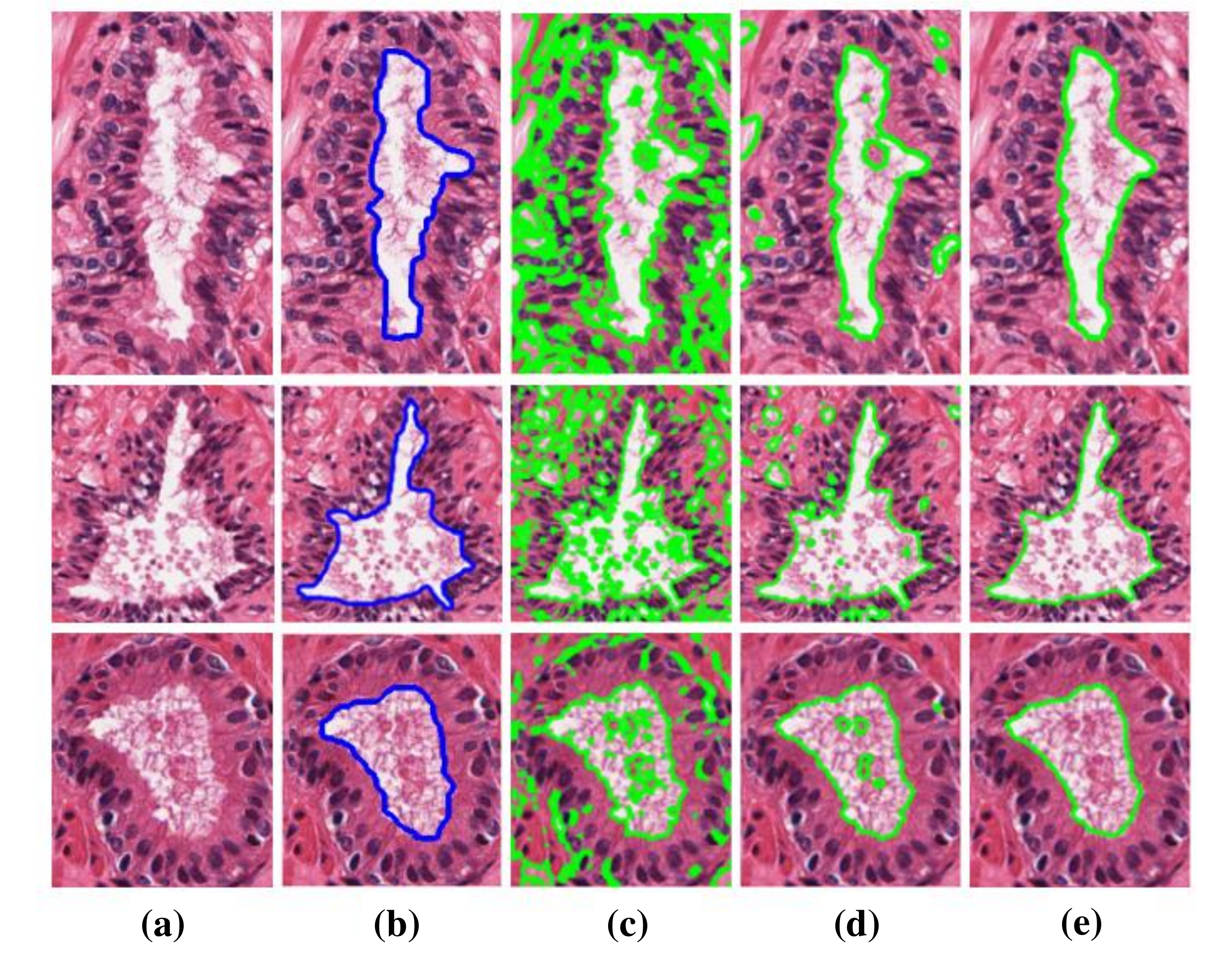}}
\caption{Method comparison using some segmentation results example for prostate cancer analysis in biopsy. This figure corresponds to Fig.2 in \cite{xu2010markov}.}
\label{fig:RAC}
\end{figure}

The research team also proposes literature focusing on automated segmentation methods. In~\cite{xu2010markov}, an MRF driven region-based active contour model (MaRACel) is presented for segmentation tasks. One limitation of most RAC models is the assumption that images of each spatial location is statistically independent. Considering that fact, the MRF prior is combined with the AC model to consider the relationship between different spatial location information. The results tested on 200 images from prostate biopsy samples show that the average sensitivity, specificity and positive predictive value were 71\%, 95\%, and 74\%, respectively. Fig.~\ref{fig:RAC} shows segmentation results comparison between the proposed method and other methods.

\begin{figure}[htbp]
  \centering
  \centerline{\includegraphics[width=0.98\linewidth]{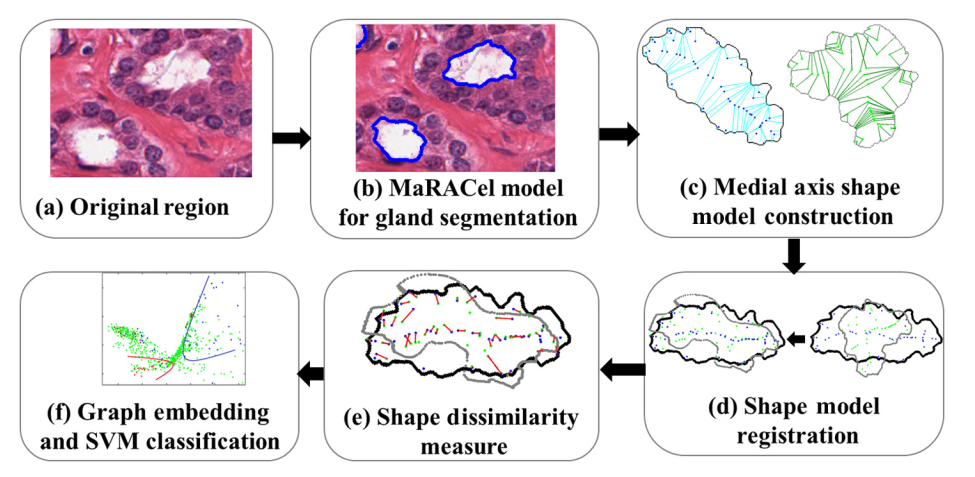}}
\caption{The work-flow of MaRACel model for gland segmentation as well as explicit shape descriptors for Gleason grading. This figure corresponds to Fig.1 in~\cite{XuConnecting}.}
\label{fig:RAC2}
\end{figure}

In~\cite{XuConnecting}, the preliminary work is extended. The authors introduce a method for incorporating an MRF energy function into an AC energy functional --an energy functional is the continuous equivalent of a discrete energy function. The MaRACel is also tested in the task of differentiation of Gleason patterns 3 and 4 glands (the flowchart is shown in Fig.~\ref{fig:RAC2}), beside the segmentation of glands task. The proposed methodology finally gets gland segmentation Dice of 86.25\% in 216 images and Gleason grading AUC of 0.80 in 55 images obtained from 11 patient studies from the Institute of Pathology at Case Western Reserve University.

\subsubsection{Works of Other Research Teams}
For the purpose of assisting pathologists in  classifying meningioma tumors correctly, a series of texture features extraction and texture measure combination methods are introduced in~\cite{al2010texture}. A Gaussian Markov random field model (GMRF)  is employed, which contains three order Markov neighbors. Seven GMRF parameters are inferred applying the least square error estimation algorithm. The combined GMRF and run-length matrix texture measures on a set of 80 pictures are proved to outperform all other combinations (e.g. Co-occurrence matrices, fractal dimension et~\textit{al}.) considering the performance of quantitatively characterizing the meningioma tissue, obtaining an overall classification accuracy of 92.50\%.

In~\cite{sun2020hierarchical, sun2020hierarchical2, Sun2020GHIS}, a hierarchical conditional random field (HCRF) model is developed to localize abnormal (cancer) areas in gastric histopathology image. The post-processing step is 
based on the MRF and morphological operations for boundary smoothing as well as noise 
removal. The detail information of HCRF will be discussed in next section.

\subsection{Summary}
A summary of the MRF methods for pathology image analysis is illustrated in 
Table~\ref{MRFsum}, which contains some essential attributes of the summarized research papers. Each row indicates publication year, reference, research team, 
input data, disease, data preprocessing, segmentation and classification techniques, 
and the result of an individual paper. It can be concluded from the table that MRFs 
has a wide range of application in various disease detection, and mostly used in the 
diagnosis of CaP and eye disease. Histopathology images are the most common input data, 
cytopathology images followed. Among these researches, the MRF models are applied in 
segmentation and classification tasks in most cases, and they are also used for 
postprocessing in~\cite{sun2020hierarchical} and feature extraction 
in~\cite{al2010texture}. Besides, with MRF theory development, some researchers 
propose the model's improvement or variants, such as PPMM, GMRF, and Connecting MRF, 
whose modelling abilities are more expansive comparing to the classical the MRFs. 
Moreover, the feature extracted algorithm for classification is more complex over 20 years. 
Since 2015, advanced machine learning models (SVM, ANN, Random Forest et~\textit{al}.) 
have been integrated into related research serving as classifiers, and the final 
results are significantly improved in larger datasets compared to those who use 
Bayesian classifier.
\renewcommand\arraystretch{1.5}
\begin{table*}
 \caption{Summary of concluded papers for pathology image analysis using MRFs. ( Specificity (Sp), Positive predictive (PP), Overlapping ratio (Oratio), Sensitivity (Sn), Area of the ROC curve (AUC), Precision (P), Mean absolute distance (MAD), Dice (D),  Recall (R), Zijdenbos similarity index (ZSI), Accuracy (Acc), Dice Coefficient (DC), F-measure (F),  Prostate Cancer (CaP)).}
     \scriptsize
    \begin{tabular}{p{1.3cm}<{\raggedright}p{1.1cm}<{\raggedright}p{1.7cm}<{\raggedright}p{2cm}<{\raggedright}p{0.8cm}<{\raggedright}p{1.8cm}<{\raggedright}p{2cm}<{\raggedright}p{1.5cm}<{\raggedright}p{1cm}<{\raggedright}}
   \Xhline{1.2pt}
        \textbf{Year, Ref, Research team} & \textbf{Disease} & \textbf{Input data} & \textbf{Task} & \textbf{Random field type} & \textbf{Optimization techniques} & \textbf{Feature extraction} & \textbf{Classification} & \textbf{Result evaluation} \\ \Xhline{1.2pt}
        2002, \cite{meas2002color}, Meas-Yedid et \textit{al}. & -- & Mouse lung slice, 23 images & Segmentation (cell nuclei, immune cells and background), Immune cell counting & MRF & -- & -- & -- & Acc=100\%. \\ \hline
        2004, \cite{won2004segmenting}, Won et \textit{al}. & -- & Blood cell, 22 images & Segmentation (nucleus, cytoplasm, red blood cell , and background) & MRF & Deterministic relaxation & Smoothness constraint and high gray level variance & -- & Acc=100\%. \\ \hline
        2008, \cite{monaco2008detection}, Monaco et \textit{al}. & CaP  & 4 WMHSs & Identification and segmentation of regions of CaP & MRF & ICM & Square root of gland area  & Bayesian classification & Sn=86.7\%, Sp=95.24\%. \\ \hline
        2009, \cite{monaco2009weighted}, Monaco et \textit{al}. & CaP  & 20 WMHSs & Identification and segmentation of regions of CaP & PPMM & WICM & Square root of gland area  & Bayesian classification & -- \\ \hline
        2009, \cite{monaco2009probabilistic}, Monaco et \textit{al}. & CaP  & 20 WMHSs & Identification and segmentation of regions of CaP & PPMM & WICM & Square root of gland area  & Bayesian classification & Sn=77\%, Sp=82\%,  AUC=0.87. \\ \hline
        2009, \cite{zou2009Fuzzy}, Zou et \textit{al}. & -- & 821 cells & Segmentation & MRF & FCM with PSO & -- & -- & Acc=86.45\%. \\ \hline
                2009, \cite{basavanhally2009computerized}, Basavanhally et \textit{al}. & Breast cancer  & HER2+ H\&E lymphocytes, 41 images & Identification and segmentation of regions of CaP & MRF & ICM & Voronoi diagram, Delaunay triangulation,and minimum spanning tree  & Bayesian classifier, SVM & Acc=90.41\%. \\ \hline
        2010, \cite{monaco2010high}, Monaco et \textit{al}. & CaP  & 40 WMHSs & Identification and segmentation of regions of CaP & PPMM & WICM & Square root of gland area  & Bayesian classification & Sn=87\%, Sp=90\%. \\ \hline
        2010, \cite{al2010texture}, Al-Kadi et \textit{al}. & Meningioma  & Meningioma tumour,  80 images & Classification (malignant or benign) & GMRF & Least square error estimation & Fractal dimension, grey level co-occurrence matrix,  grey level run-length matrix, GMRF & Bayesian classification & Acc=92.50\%. \\ \hline
        2010, \cite{xu2010markov}, Xu et \textit{al}. & CaP  & 200 prostate biopsy needle images & Segmentation of prostatic acini  & MRF & -- & -- & -- & Sn=71\%, Sp=95\%, PP=74\%. \\ \hline
        2011, \cite{yu2011detection}, Yu et \textit{al}. & CaP  & 27 WMHSs & Identification and segmentation of regions of CaP & PPMM & ICM & CFD & Bayesian Classification & AUC=0.831. \\ \hline
        2011, \cite{monaco2011weighted}, Monaco et \textit{al}. & CaP  & 27 WMHSs & Identification and segmentation of regions of CaP & MRF & E-MCMC, M-MCMC & Gland area  & WMPM classification & -- \\ \hline
        2014, \cite{bioucas2014alternating}, Bioucas-Dias et \textit{al}. & Teratoma  & Teratoma tissue, a 1600 × 1200 image & Classification of four categories of teratoma tissues & HMRF & EM based algorithm & -- & -- & Acc=84\%. \\ \hline
        2014, \cite{SalazarSegmentation}, Salazar-Gonzalez et \textit{al}. & Eye disease  & Fundus retinal images, DIARETDB1, DRIVE,  STARE public dataset & Segmenting of blood vessel and optic disk & MRF & Max-flow algorithm & -- & -- & Oratio=0.8240, MAD=3.39, Sn=98.19\% (in DRIVE). \\  \Xhline{1.2pt}
    \end{tabular}
    \label{MRFsum}
\end{table*}

\renewcommand\arraystretch{1.5}
\begin{table*}
     \scriptsize
    \begin{tabular}{p{1.3cm}<{\raggedright}p{1.1cm}<{\raggedright}p{1.7cm}<{\raggedright}p{2cm}<{\raggedright}p{0.8cm}<{\raggedright}p{1.8cm}<{\raggedright}p{2cm}<{\raggedright}p{1.5cm}<{\raggedright}p{1cm}<{\raggedright}}
   \Xhline{1.2pt}
        \textbf{Year, Ref, Research team} & \textbf{Disease} & \textbf{Input data} & \textbf{Task} & \textbf{Random field type} & \textbf{Optimization techniques} & \textbf{Feature extraction} & \textbf{Classification} & \textbf{Result evaluation} \\ \Xhline{1.2pt}
          2015, \cite{liu2015study}, Liu et \textit{al}. & Melanoma & Melanoma in Demoscopy, 1039 images &  Lesion region segmentation and classification (malignant or benign) & MRF & -- & Symmetry, size, shape, maximum diameter, Gray level co-occurrence matrix, color features, SIFT & SVM classfier & Acc=94.49\%, Sn=95.67\%, Sp=94.31\%.  \\ \hline
        2016, \cite{paramanandam2016automated}, Paramanandam et \textit{al}. & Breast cancer & High-grade breast cancer images, 512 nuclei & Segmentation of the individual nuclei & MRF & Loopy Back Propagation & Tensor voting method & -- & P=96.57\%, R=74.80\%,  DC= 88.3\%. \\ \hline
        2016, \cite{Razieh2016Microaneurysms}, Razieh et \textit{al}. &  Microane- -urysms & Fundus retinal images, DIARETDB1  public dataset & Microaneurysms segmentation & MRF & Simulated annealing & Shape-based features, Intensity and color based features, Features based on Gaussian distribution of MAs intensity & SVM classfier & Sn=82\%. \\ 
 \hline
 2016, \cite{zhao2016automatic}, Zhao et \textit{al}. & Cervical cancer & Color cervical smear images, Herlev and real-world datasets & Segmentation of cytoplasm and nuclei & MRF & Iterative adaptive classified algorithm & Pixel intensities and the shape of superpixel patches & -- &  Herlev: ZSI=0.93, 0.82; real-world: ZSI=0.72, 0.71 (cytoplasm, nuclei). \\ 
 \hline
        2017, \cite{mungle2017mrf}, Mungle et \textit{al}. & Breast cancer & Breast cancer immunohistochemical images,  65 patients’ tissue slides & ER scoring & MRF & EM, ICM and Gibbs sampler (with simulated annealing) & R, G and B values of individual cell blobs & ANN & F=96.26\%. \\ \hline
        2017, \cite{XuConnecting}, Xu et \textit{al}. & CaP  & 600 prostate biopsy needle images & Segmentation of prostatic acini and Gleason grading & Connecting MRFs & -- & \cite{Sparks2013Explicit} & SVM classfier & Segmentation: D=86.25\%, Gleason grading: AUC= 0.80. \\ \hline
        2018, \cite{dholey2018combining}, Dholey et \textit{al}. & Lung Cancer & Papanicolaou-stained cell cytology,  600 image & Segmentation of the nucleus and classification (Small Cell Lung Cancer and Non-small Cell Lung Cancer) & GMM-HMRF & HMRF-EM & SIFT, $K$-Means Clustering, Construction of Visual Dictionary & Random Forest Training (Bag-of-Words) & Sn= 98.88\%, Sp=97.93\%. \\ \hline
        2019, \cite{su2019cell}, Su et \textit{al}. & -- & Bone marrow smear, 61 images & Segmentation & HMRF & EM & Color intensity & $k$-means & Acc=96.85\%.  \\ 
        \Xhline{1.2pt}
    \end{tabular}
\end{table*}
\section{CRFs for Pathology Image Analysis}
\label{sec:CRFs}
In this section, unlike the MRFs concluded before, it is known that the CRFs take 
segmentation and classification tasks simultaneously. Instead, these researches are 
categorized into microscopic images and micro-alike (close-up/macro images) two images 
of analysis work by the input dataset's property.

Their main differences can be concluded into two points: magnification and application 
scenarios. Most of the work to obtain micro-alike image can be accomplished within the 
magnification range of $2 \times$ to $15 \times$, such as endoscopy and ophthalmoscopy. 
On the contrary, $20 \times$ and $40 \times$ optical magnifications are most frequently 
used for acquiring microscopic images, such as examining tissues and/or cells under a 
microscope for cancer diagnosis. Due to their different property, they are used in 
different application scenarios. Take a colposcope as an example. A lower magnification yields a broader horizon as well as greater depth of field for the cervix examination.
More magnification is not necessarily better, since there are certain trade-offs as magnification increases: the field of view becomes smaller, the depth of focus diminishes, and the illumination requirement increases~\cite{WHOcolposcopy}.  When examining cells and tissues removed from suspicious ‘lumps and bumps’, and identifying whether they are 
from tumor or normal tissue, a microscope of high magnification is indispensable.

The related research papers are concluded first. Finally, a summary of method analysis is given in the last paragraph.

\subsection{Microscopic Image Analysis}
In~\cite{Xu2009Conditional}, a method based on multi-spectral data is proposed for cell segmentation. A CRF model incorporating spectral data during inference is developed. 
The loopy belief propagation algorithm is applied to calculate the marginal distribution, 
which also solves the optimal label configuration problem.  The proposed CRF model achieves 
better results because the spectral information describing the relationship between 
neighboring bands helps to integrate spatial and spectral constraints within the 
segmentation process. 12 FNA samples are used for testing, and the result shows that the 
CRF model could help to get over segmentation difficulties when the contrast-to-noise 
ratio is poor.

In~\cite{rajapakse2011staging}, a method is proposed for identifying disease states by 
classifying cells into different categories. Single cell classification can be divided into three steps: Firstly, level sets and marker-controlled watershed algorithm are applied for cell segmentation. Second, wavelet packets is employed to extract the the cell feature. Finally, SVM and CRF are served as the classifiers. Image information is represented by CRF to characterize the features and distribution of adjacent cells, which plays an significant role in tissue cells classification task. Its potential is related to the output discriminant value of SVM. After initialization of the CRF, considering that most of cells do not regularly distribute over the tissue, an algorithm for determining the optimal graph structure on the basis of local connectivity is proposed. This method is tested in lung tissue images 
containing 9551 cells, yielding specificity $96.52\%$, sensitivity $48.30\%$, and accuracy 
$90.26\%$.

\begin{figure}[htbp]
  \centering
  \centerline{\includegraphics[width=1.1\linewidth]{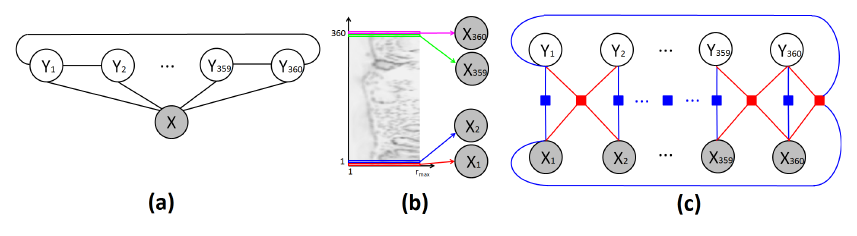}}
\caption{A circular area with radius of $r_{max}$ is converted to a fixed size polar image of 360 $rows$ $\times$  $r_{max}-columns$ after the transformation mentioned in the article.  This figure corresponds to Fig.2 in~\cite{fu2012glandvision}.}
\label{fig:polar1}
\end{figure}

\begin{figure}[htbp]
  \centering
  \centerline{\includegraphics[width=1.1\linewidth]{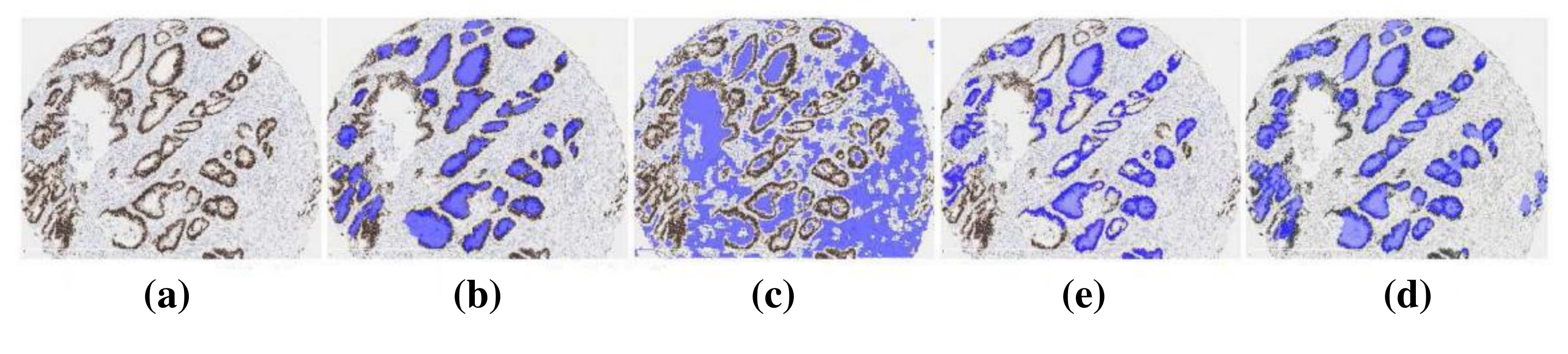}}
\caption{Segmentation results of different approaches. The detailed information of this figure can be seen in  Fig.20 in~\cite{fu2012glandvision}.}
\label{fig:polar2}
\end{figure}

In~\cite{fu2012glandvision}, a novel method is presented for glandular structures detection using microscopic images of human colon tissues, where the images are transformed to polar space first, then a CRF model (shown in Fig.~\ref{fig:polar1}) is introduced to find the gland's boundary of and a support vector regression (SVR) based on visual characteristics is proposed to evaluate whether the inferred contour corresponds to a real gland. In the final step, the results of the previous step combined with these two methods are utilized to find the GlandVision algorithm ranking of all candidate contours, and then generate the final result. In the inference process of the CRF, two chain structures are applied to approximate this circulate graph, which uses Viterbi algorithm to find the optimal results. The authors use the combination of cumulative edge map and the original polar image to generate the unary potential and the Gaussian edge potential as the pairwise potential. Besides, the thresholding algorithm is applied to eliminate most of the false positives regions generated in this step. Finally, a segmentation accuracy of 0.732 is achieved on the dataset containing 20 microscopic images of human colon tissues for training and testing (shown in Fig.~\ref{fig:polar2}). Based on the work above, in~\cite{fu2014novel}, a task called inter-image learning is introduced, which predicts whether those sub-images containing glands. A linear SVM is applied to tackle this problem, using the sum of the node potential of the CRF in contour detection task along with other mid-level features (listed in Table \ref{CRFsum} in detail). This research also finds that all the gland contours proposed by the random field which rank according to the learned SVM score achieves the best result compared to other index, so it is used as the final output of the CRF. Based on the same dataset, this method gets an accuracy of 80.4\%. Beside the grey-scale images, it is also tested in 24 H\&E stained images, yielding sensitivity 82.35\%, specificity 93.08\%, accuracy 87.02\% and Dice 87.53\%.

A system is presented to segment necrotic regions from normal regions in brain pathology images based on a sliding window classification method followed by a CRF smoothing in~\cite{manivannan2014brain}. First, four features are extracted and encoded by Bag-of-words (BoW) algorithm. Then, an SVM classier is applied in the sliding windows using the features extracted in the last step. Thirdly, the CRF model is applied to discard noisy isolated predictions and obtain the final segmentation with smooth boundaries. The node and edge potentials of the CRF is defined using the probability map provided by SVM. In 35 training data provided by the MICCAI 2014 digital pathology challenge, the proposed method gets a segmentation accuracy value of 0.66.

In~\cite{wang2016deep}, an end-to-end algorithm is proposed based on fully convolutional networks (FCN) for inflammatory bowel disease diagnosis to recognize muscle and messy areas. For purpose of incorporating multi-scale information into the model, a specific field of view (FOV) method is applied. 
The structure of multi-scale FCN is as follows: First, different FCNs are used to make each FCNs process the input image with a different FOV; then, the results of these FCNs are merged to form a score map; finally, the fusion score map is computed through the soft-max function for  the cross-entropy classification loss calculation. The CRFs are severed as post-processing method of FCN. It takes the probability generated by the FCN as the unary cost, and also considers the pairwise costs, so that the results are smooth and consistent. Tested in 200 whole slides images of H\&E stained histopathology tissue, the authors manage to achieve an accuracy of 90\%, region intersection over union (IU) of 56\%.

The degree of deterioration of breast cancer is highly related to the number of mitoses in a given area of the pathological image. Considering that fact, a multi-level feature hybrid FCNN connecting a CRF mitosis detection model is proposed in~\cite{wu2017STUDY}. On the open source ICPR MITOSIS 2014 dataset, the proposed classification methodology is found to have F-score 0.437.

In \cite{he2017Reasearch}, a cell image sequence morphology classification method based on linear-chain condition random field (LCRF) is presented. Firstly, this problem is modelled as a multi-class classifier based on LCRF, a conditional probability distribution model assuming that $X$ and $Y$ own identical structure. Then, a series of features are extracted to describe the internal motion for cells image sequence, including deformation factor and dynamic texture. Lastly, the model parameter is estimated by discrimination learning algorithm called margin maximization estimation and the image sequence classification result is produced according to the input feature vectors. The effectiveness of the model is verified on micro image sequence data set of pluripotent stem cells from University of California, Riverside, USA, yielding an accuracy value of 0.9346.

\begin{figure}[htbp]
  \centering
  \centerline{\includegraphics[width=0.98\linewidth]{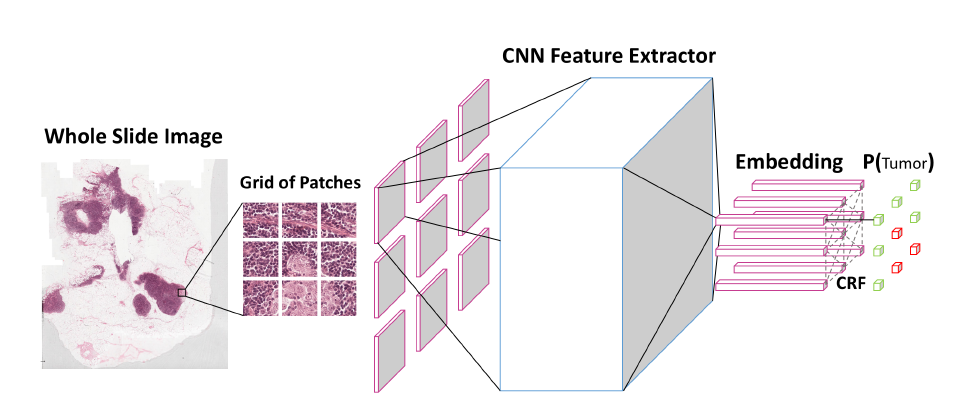}}
\caption{The architecture of NCRF model. This figure corresponds to Fig.1 in~\cite{li2018cancer}.}
\label{fig:NCRF}
\end{figure}

In~\cite{li2018cancer}, a neural conditional random field (NCRF) DL framework is proposed to detect cancer metastasis in WSIs. The NCRF is directly incorporated on top of a CNN feature extractor (called ResNet) forming the whole end-to-end algorithm. Fig.~\ref{fig:NCRF} illustrates the overall architecture of the NCRF. Specifically, the authors use the mean-field inference to approximate marginal distribution of each patch label. Then, after the network calculates the cross-entropy loss, the entire model is trained using the back propagation algorithm. On the Camelyon16 dataset, including 270 WSIs for training, 130 tumor WSIs for testing, an average free response receiver operating characteristic (FROC) score of 0.8096 is achieved finally.

To recognize cancer regions of pathological slices of gastric cancer, a reiterative learning framework is proposed in~\cite{liang2018Weakly}, which first extracts the region of interest in diagnosis process, then the patch-based FCN is trained, and finally uses the FC-CRF to perform overlap region prediction and post-processing operations. However, the performance of the CRF in their task is not satisfactory, because several erroneous data distributions in such weak annotation task may lead to bad performance. After adding the CRF to the model, a mean intersection over union Coefficient (IoU) value in test set decreases from 85.51\% to 84.85\%. Based on the work above, this research team improves the network structure, proposing a deeper segmentation algorithm deeper U-Net (DU-Net) in~\cite{liang2018deep}. In this case, post-processed with the CRF is proved to boost the performance and the IoU value increases to 88.4\% in the same dataset. 

\begin{figure}[htbp]
  \centering
  \centerline{\includegraphics[width=0.98\linewidth]{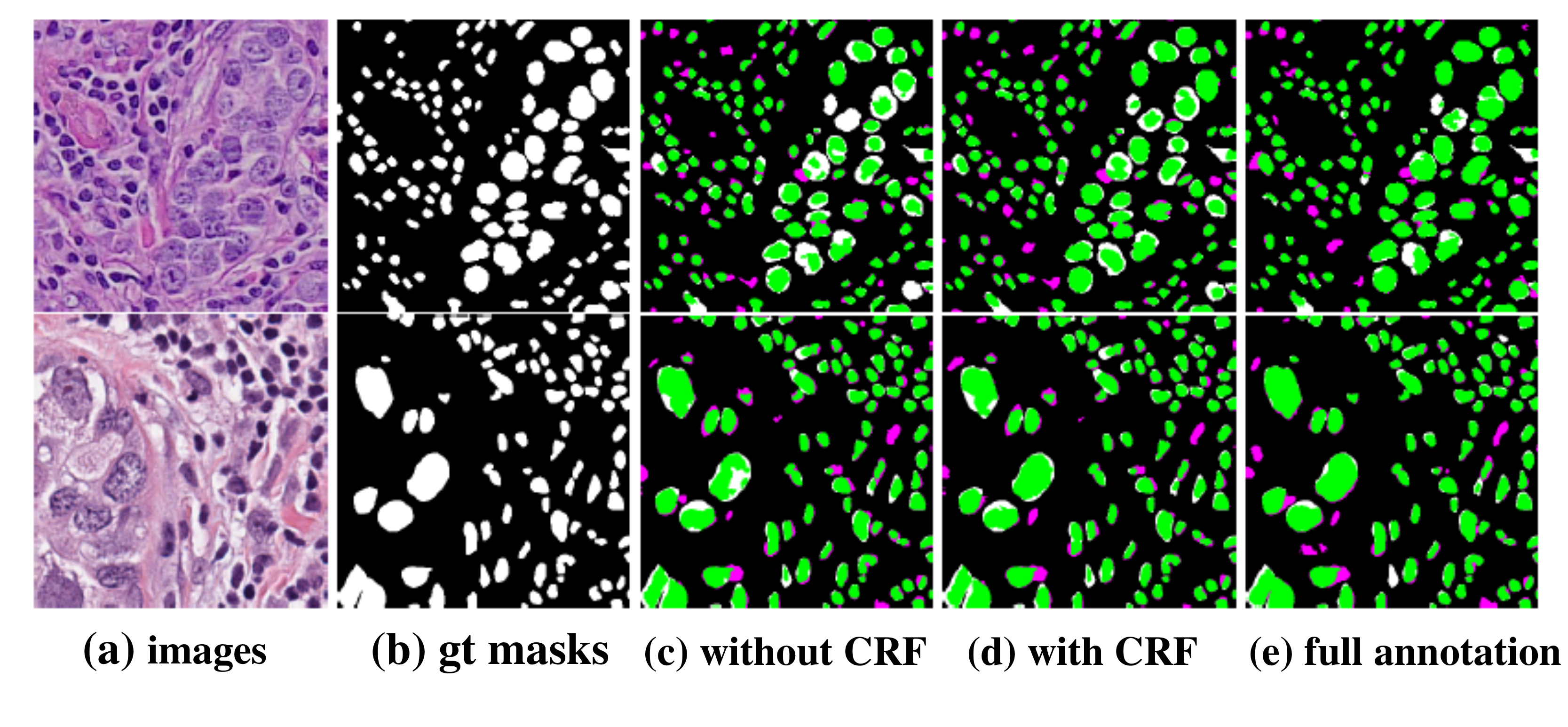}}
\caption{Comparison of weakly and fully supervised training. This figure corresponds to Fig.4 in \cite{qu2019weakly}.}
\label{fig:weak}
\end{figure}

A method using weak annotations for nuclei segmentation is proposed in~\cite{qu2019weakly}. Firstly, in order to derive complementary information, the Voronoi label and cluster label which serve as the in initial label are generated with the points annotation image. Second, label produced in last step is employed to train a CNN model. Dense CRFs is embedded into the loss function to further refine the model. In the experiment, lung cancer and MultiOrgan dataset are used for testing, both achieving accuracy over 98\%. The result is shown in Fig.~\ref{fig:weak}.

A cell segmentation method with texture feature and spatial information is developed in~\cite{sara2019use}. In the first step, features are extracted and utilized to train machine learning model, providing pre-segmentation result. In the second step, the image is post-processed by MRF and CRF model for binary denoising. The proposed segmentation methodology obtains F-score, Kappa and overall accuracy of 86.07\%, 80.28\% and 91.79\%, respectively. 

\begin{figure}[h]
  \centering
  \centerline{\includegraphics[width=0.98\linewidth]{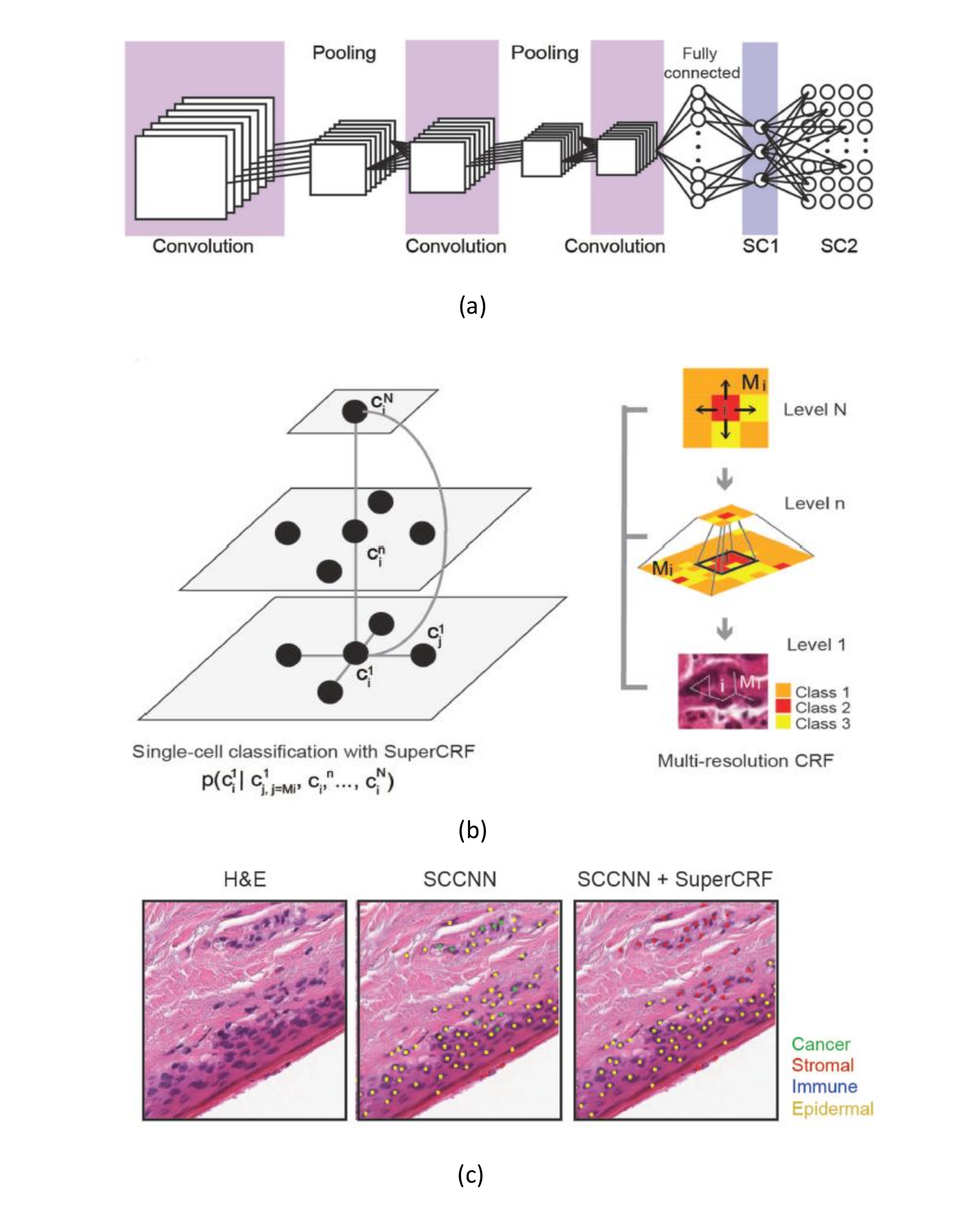}}
\caption{Overview of the SuperCRF framework for melanoma image analysis. This figure corresponds to Fig.1 in \cite{Zormpas2019Superpixel}.}
\label{fig:super}
\end{figure}

Inspired by the way pathologists perceive the regional structure of tissue, a new multi-resolution hierarchical network called SuperCRF is introduced in~\cite{Zormpas2019Superpixel} to improve cell classification. A CNN that considers spatial constraint is trained for WSIs detection and classification. Fig.~\ref{fig:super}(a) illustrates this network clearly. Then, a CRF, whose architecture is shown in Fig.~\ref{fig:super}(b), is trained by combining the adjacent cell with tumor region classification results based on the super-pixel machine learning network. Subsequently, the labels of the CRF nodes representing single-cell are connected to the regional classification results from superpixels producing the final result. Segmentation accuracy, precision and recall of 96.48\%, 96.44\%, and 96.29\% are achieved on 105 images of melanoma skin cancer, shown in Fig.~\ref{fig:super}(c).

\begin{figure}[h]
  \centering
  \centerline{\includegraphics[width=0.98\linewidth]{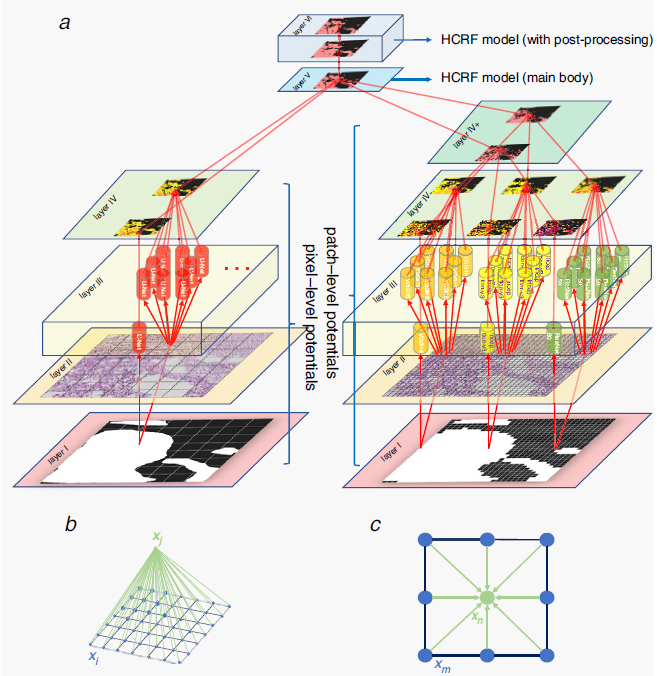}}
\caption{Overview of HCRF framework for gastric histopathological image analysis. This figure corresponds to Fig.2 in~\cite{sun2020hierarchical}.}
\label{fig:HCRF}
\end{figure}

In~\cite{sun2020hierarchical, sun2020hierarchical2, Sun2020GHIS}, a gastric histopathology image segmentation method based on HCRF is introduced for abnormal regions detection. The structure of the HCRF model is presented in Fig.~\ref{fig:HCRF}(a). Firstly, a DL network U-Net is retrained to construct pixel-level potentials. Meanwhile, for purpose of building up patch-level potentials, the authors fine tune another three CNNs, including ResNet-50, VGG-16, and Inception-V3. Adding pixel-level and patch-level potentials contribute to integrate information in different scales. Besides, binary potentials of their surrounding image patches are formulated according to the ‘lattice’ structure described in Fig.~\ref{fig:HCRF}(b)(c). When the HCRF model is constructed, the post-processing operation is finally adopted to refine the results. Researchers test the performance of HCRF on a public gastric histopathological image dataset containing 560 images, achieving 78.91\% segmentation accuracy.

In \cite{li2020automated}, a CNN-based automatic diagnosis method for prostate cancer images is proposed. Adopted the core concept of a CNN using multi-scale parallel branch, a multi-scale standard convolution method is developed. An architecture that combines the atrous spatial pyramid pooling from Deeplab-V3 and the network mentioned above as well as their feature maps are cascaded together to obtain initial segmentation results. Subsequently, a  post-processing method that is based on CRF model is adopted to the prediction. The proposed system yielded an mIOU and overall pixel accuracy value of 0.773 and 0.895, respectively, for Gleason patterns segmentation.

\begin{figure}[htbp]
  \centering
  \centerline{\includegraphics[width=0.98\linewidth]{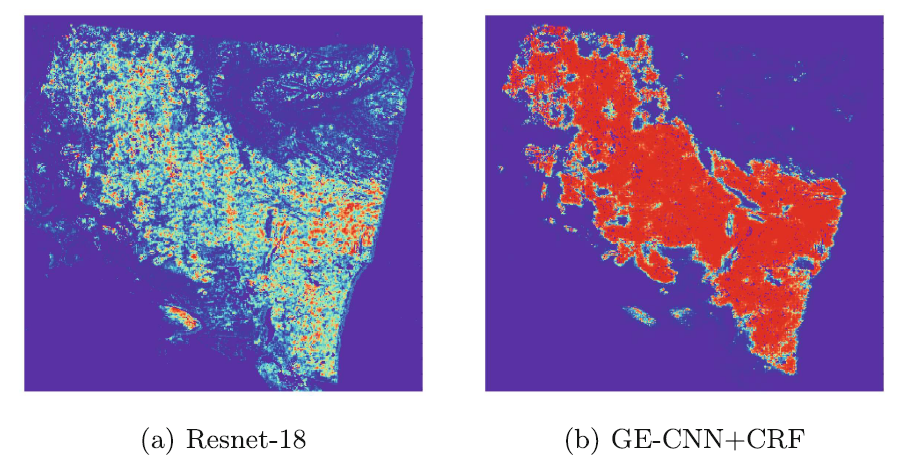}}
\caption{The probabilistic heat maps comparison between Resnet-18 and GECNN-CRF. This figure corresponds to Fig.7 in~\cite{Dong2020GECNN}.}
\label{fig:GECNN}
\end{figure}

In order to enhance the effectiveness of the network and solve the limitation of inconsistent data characteristics caused by random rotation in the learning process, a CNN based on a special convolution method and CRF is developed in~\cite{Dong2020GECNN}. 
This method uses the output probability of the CNN model to establish the unary potential of FC-CRFs. And the feature map of adjacent patches is used to design a pair-wise loss function to represent the relationship between two blocks. The mean field approximate inference is applied to acquire the marginal distribution and calculate the cross entropy loss of the model. Thus, the whole network can be trained end-to-end. The experiment verifies that applying the CRFs can obviously improve the performance. Fig.~\ref{fig:GECNN} shows the comparison of the probability heat map of the cancer regions. And a segmentation accuracy and AUC of 85.1\% and 91.0\% are finally achieved on the test set.

\subsection{Micro-alike Image Analysis}
\begin{figure}[htbp]
  \centering
  \centerline{\includegraphics[width=0.98\linewidth]{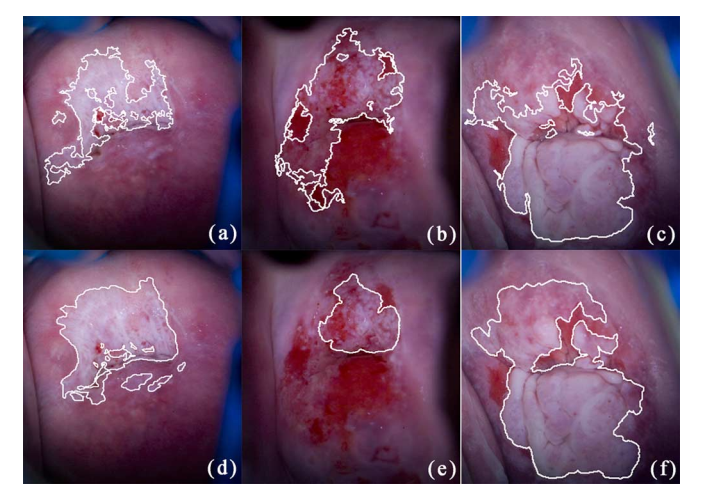}}
\caption{Acetowhite region detection by proposed algorithm. This figure corresponds to Fig.9 in ~\cite{park2011domain}.}
\label{fig:cervical2}
\end{figure}

An intelligent semantic image analysis method is developed to detect cervical cancerous lesion based on colposcopy images in~\cite{park2010semantic}.  First, preprocessed images semantics maps are generated. $k$-means clustering is then applied to image segmentation. Thirdly, some distinguished features of cervical tissues are extracted. In the final step, a CRF-based method classifies the tissue in each region (segmented in the second step) as normal or abnormal. The CRF-based classifier combines the outputs of neighboring regions produced by $k$-NN and LDA classifier in a probabilistic manner. In the experiment of detecting neoplastic areas on colposcopic images obtained from 48 patients, an average sensitivity of 70\% and a specificity of 80\% are achieved. As an extension of this work, the performance evaluation step is improved in~\cite{park2011domain}. 
It is of great significance for the clinic to detect the abnormal area and provide a high-accuracy diagnosis. Considering this fact, a window-based method is proposed. The method first divides the image into patches, and then compares their classification results according to the ground truth in the histopathology images. In the same dataset, compared with expert colposcopy annotations (AUC=0.7177), the proposed method gains better performance (AUC=0.8012). Fig.~\ref{fig:cervical2} presents examples of abnormal areas detected by proposed algorithm comparing the diagnostic accuracy with that of the colposcopist.

\begin{figure}[htbp]
  \centering
  \centerline{\includegraphics[width=0.98\linewidth]{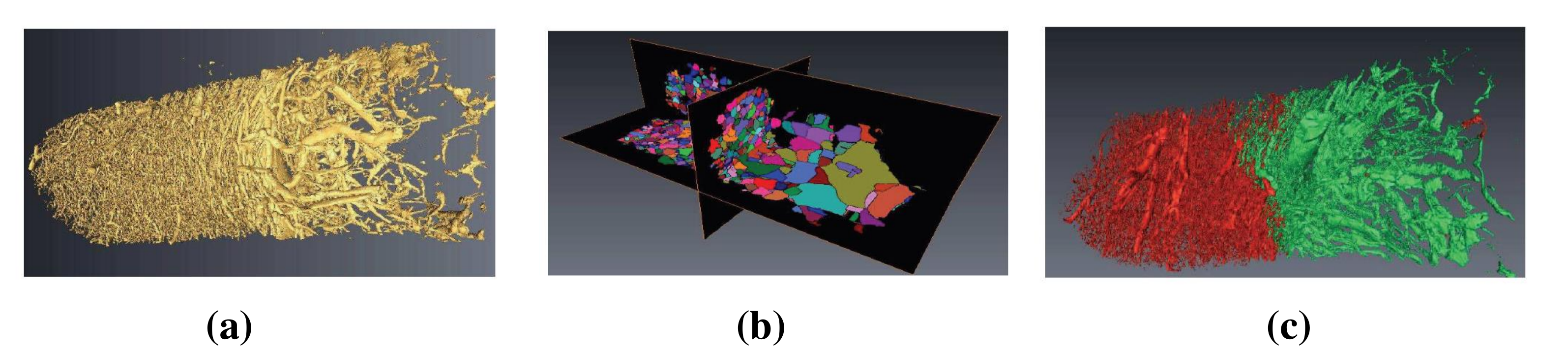}}
\caption{(a)Full data rendering after merging steps; (b) Watershed on the distance map; (c) The illustration of the segmentation result: the area in red represents normal tissue and green represents tumor. This figure corresponds to Fig.1,3,6 in original paper~\cite{Xavier2011Brain}.}
\label{fig:brain}
\end{figure}

In~\cite{Xavier2011Brain}, a novel framework is presented for dividing vascular network into abnormal and normal areas from only taking the  3D structure of micro-vessel into consideration, where a distance map in preprocessed images is generated by computing the Euclidean distance first. Then, the inverse watershed of the distance map  is computed. Finally, a CRF model is introduced to divide the watershed regions into tumor and non-tumor regions. Key images in the whole process are shown in Fig.~\ref{fig:brain}. This method is tested in real intra-cortical images by Synchrotron tomography, which corresponds to the experts’ expectation.

In ~\cite{Orlando2014Learning}, a method of fundus blood vessel segmentation based on discriminatively trained FC-CRF model is proposed. Unlike traditional CRF, in FC-CRF, each node is assumed to be each other's neighbors. Utilizing this method, not only the neighboring information can be considered, but also the relationship between distant pixels. Firstly, some features (Gabor wavelets, line detectors et \textit{al}.) are extracted, serving as the parameters of unary and pairwise energy combined with linear combination weight. Then, a supervised algorithm called Structured Output SVM is applied,which can learn the parameters for unary and pairwise potentials. On the DRIVE dataset, this method yields sensitivity 78.5\%, specificity 96.7\%. As an extension of this work, an intelligent method based on the similar workflow is introduced in~\cite{Orlando2016Discriminatively} to overcome a limitation of previous research: the configuration of the pairwise potentials of the FC-CRF is affected by image resolution, because it is associated with the relative distance of each pixel. 
In order to solve this problem, an optimal parameter estimation method based on the feature parameters on a single data set is proposed, and it is multiplied by the compensation factor for adjustment.
 The authors manage to achieve over 96\% accuracy and over 72\% on DRIVE, STARE, HRF and CHASEDB1 four datasets.

In~\cite{Fu2016Retinal}, for the objective of improving the accuracy of retinal vessel segmentation, a DL architecture is proposed. Fully CNNs are utilized for vessel probability map generation. Afterwards, an FC-CRF model is applied to build the long-range correlations between pixels to refine the segmentation result. Some of the unary and pairwise energy components are obtained by the vessel probability map. In the experiment, public dataset DRIVE and STARE achieve the segmentation accuracy of 94.70\% and 95.45\%. Furthermore, an improved system is introduced in~\cite{fu2016deepvessel}. 
A comprehensive deep network called DeepVessel is proposed, which consists of CNN stages and a CRF stage. A CNN integrating multi-scale and multi-level information with a delicately designed output layer to learn rich level representations is applied.
The FC-CRF is applied in the last layer of the network, which is reformulated as an RNN layer so that it can be adopted in the end-to-end DL architecture. Its unary and pairwise terms are determined by the previous layers and its loss function is combined with the CNN layer loss function minimized by standard stochastic gradient descent. DeepVessel is tested in three public databases (DRIVE, STARE, and CHASE DB1), yielding an accuracy value of 95.23\%, 95.85\% and 94.89\%, respectively. The results comparison is shown in Fig.~\ref{fig:brain}.

\begin{figure}[htbp]
  \centering
  \centerline{\includegraphics[width=0.98\linewidth]{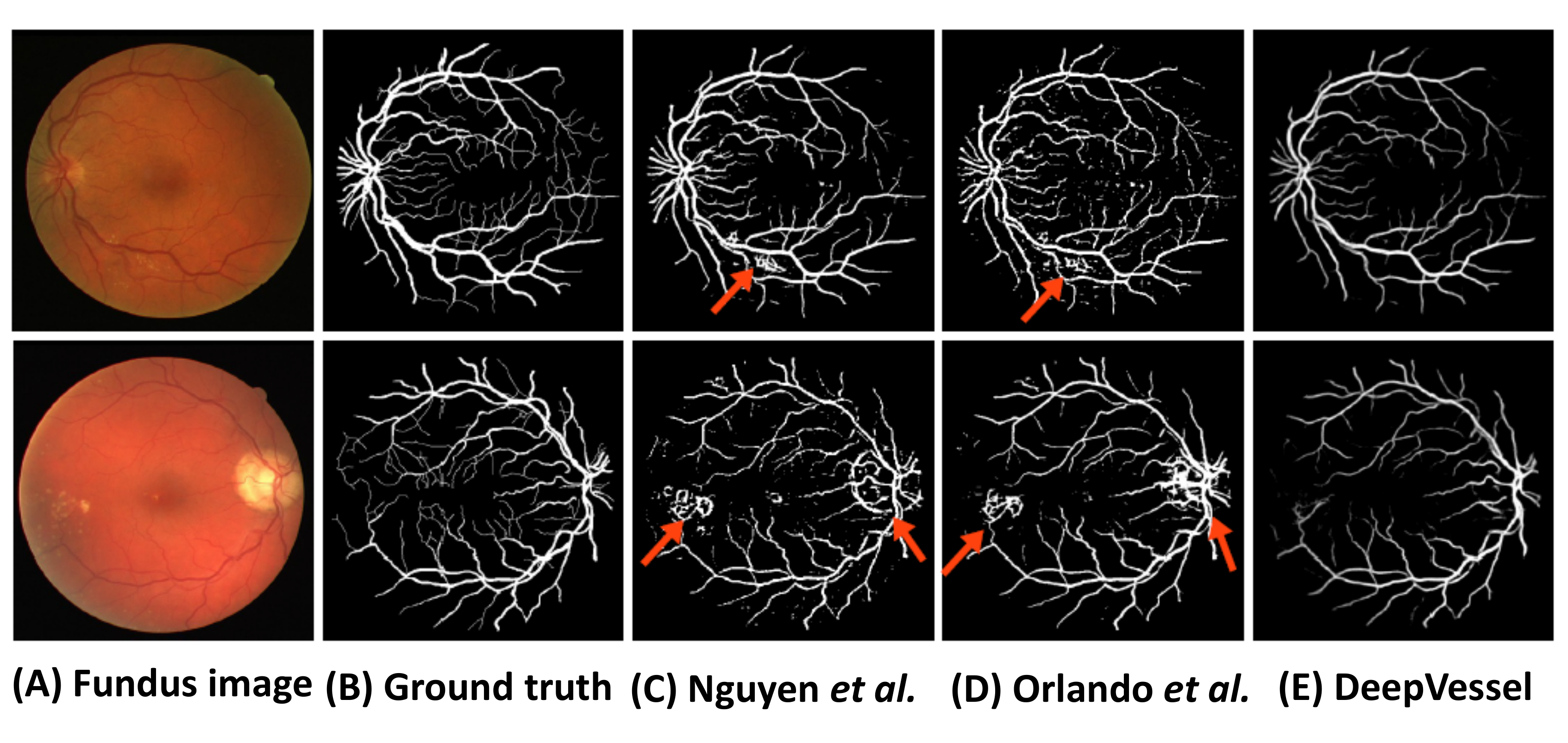}}
\caption{Retinal vessel segmentation results. This figure corresponds to Fig.1 in \cite{fu2016deepvessel}.}
\label{fig:vessel}
\end{figure}

\begin{figure*}[htbp]
  \centering
  \centerline{\includegraphics[width=0.98\linewidth]{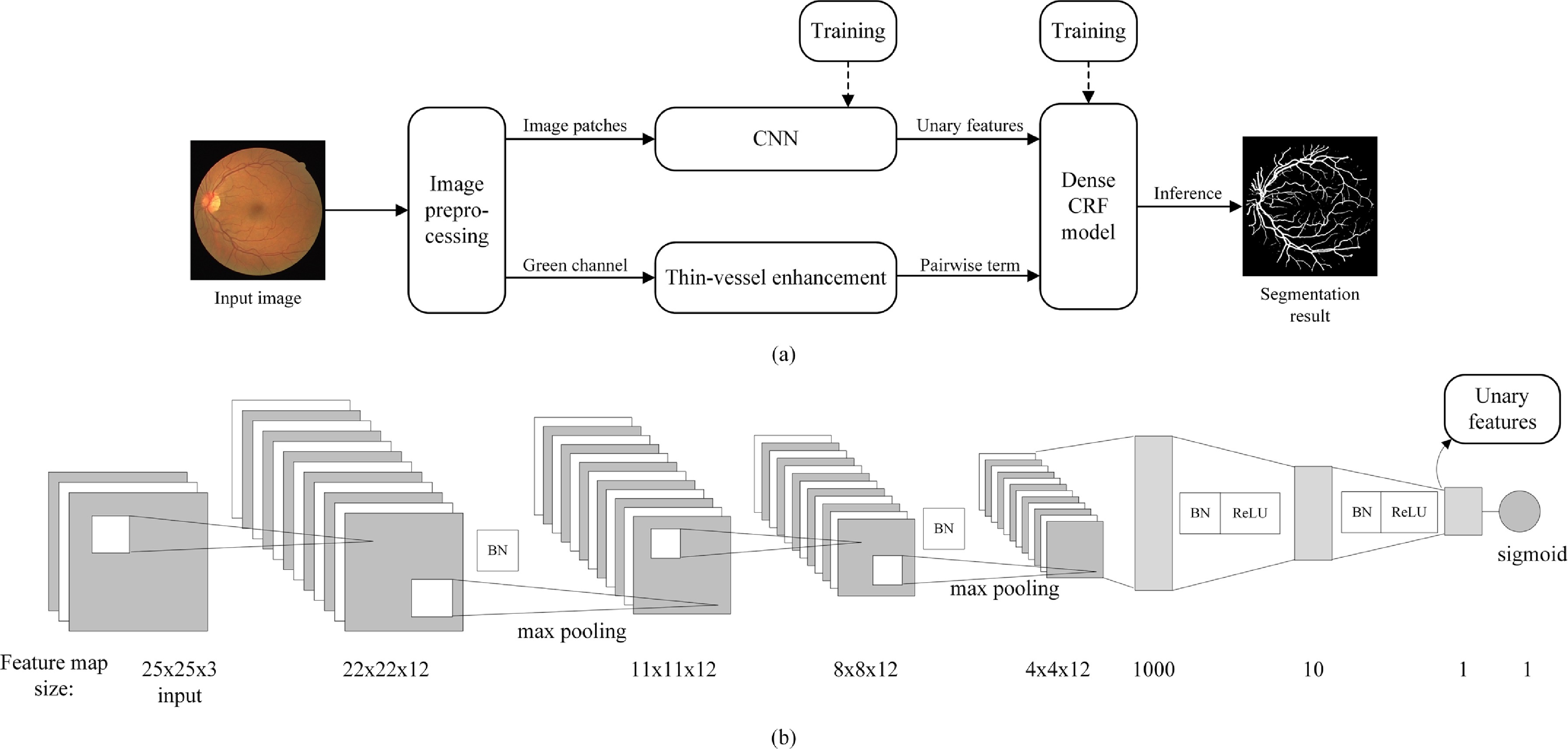}}
\caption{Overview of the retinal vessel segmentation method. (a) Work-flow of the proposed method. (b) The CNN's structure containing BN–Batch Normalization layer and ReLU–Rectified Linear Units applied for discriminative feature learning. This figure corresponds to Fig.1 in~\cite{Zhou2017Improving}.}
\label{fig:vesselfc}
\end{figure*}

A four-step method is developed for retinal vessel segmentation in \cite{Zhou2017Improving}. 
Firstly, the image is preprocessed to remove noisy edges and normalized. Then, training a CNN to generate features that can be distinguished for the linear model. Third, in order to reduce the intensity difference between narrow blood vessels and wide blood vessels, a combined filter is used on the green channel to enhance the narrow blood vessels. Finally, the dense CRF model is applied to obtain the final segmentation of retinal blood vessels, whose unary potentials are formulated by the discriminative features and pairwise potentials are comprised of the intensity value of pixel thin-vessel enhanced image. The work-flow of the whole process is shown in Fig.~\ref{fig:vesselfc}. Among DRIVE, STARE, CHASEDB1 and HRF four public dataset, proposed method achieves the best result in DRIVE (F1-score = 0.7942, Matthews correlation coefficient = 0.7656, G-mean = 0.8835).

An end-to-end algorithm is proposed in~\cite{Cl2018Multitask} for red and bright retinal lesions segmentation, which are essential biomarkers of DR. 
First, a patch-based multi-task learning framework is trained, which extends the function of U-Net. Then, CRF is used as the RNN to refine the segmentation output, and the rest of the network is applied to train the parameters of the kernel function. The softmax output of each decoding module in CNN is used as the unary potential. The binary potential is represented by the weighted sum of two Gaussian kernels. However, the results showed that the model performance deteriorated after adding the CRFs, because the CRFs add tiny false positive red lesions near blood vessels. This method is finally evaluated on a publicly available DIARETDB1 database and obtains specificity value of 99.8\% and 99.9\%, for red and bright retinal lesions detection respectively.

A novel approach combining the CNN with CRFD is proposed in~\cite{Huang2018Automatic} for purpose of detecting the optic disc in retinal image. CNN constructs the first-order potential function of CRF, and the second-order potential function is constructed by linear combination of Gaussian kernels.
Lastly, in post-processing step, a regional restricts method is adopted to obtain the super-pixel area, which is used to make the connected area label consistent. Afterwards, in order to refine the results, the posterior probability mean of the super-pixel region is calculated. This method is verified on several retina databases and yields an accuracy of 100\% in DRIVE, MESSID, DIARETDB, and DRION dataset. This method is improved and applied to other tasks like retina blood vessel segmentation and retina arteriovenous blood vessel classification in~\cite{huang2018Research} by the same research team.

An automated optic disc segmentation method  is proposed in~\cite{bhatkalkar2020improving}. The input image is trained by modified U-Net or DeepLabV3 network, and the prediction is resized and fed to the CRF for training and inference. However, the CRFs are not easily GPU accelerated leading to slower performance after added. Moreover, because of the smooth ground truth masks with little sharp deformation and the already satisfying boundary produced by neural networks, CRF has very little effect on performance improvement. No matter tested in private dataset or DRIONS-DB, RIM-ONE v.3, and DRISHTI-GS public dataset, Dice coefficient values all achieve over 95\%.

A skin lesion segmentation ensemble learning framework on the basis of multiple deep convolutional neural network (DCNN) models are proposed in~\cite{qiu2020inferring}. The whole process can be divided into three steps: First, in the training phase, multiple lesion segmentation maps are obtained from various pre-trained DCNN models, and then these segmentation maps are used as the unary potential of the CRF model. Lastly, the CRF energy is minimized and yields the final prediction. In the experiment, ISIC 2017~\cite{ISBI}and PH2~\cite{PH2A} public datasets are used for testing, and a mean Dice coefficient of 94.14\% is finally achieved. 

A novel framework for automatic auxiliary diagnosis of skin cancer is introduced~\cite{adegun2020fcn}. The task of this framework is to segment and classify skin lesions. The framework consists of two parts: The first part uses the encoder-decoder FCN to learn complex and uneven skin lesion features. Then the post-processing CRF module optimizes the output of the network. This module adopts the linear combination of Gaussian kernels as a binary potential for the objective of contour refinement and lesion boundary location. In the second part, the segmentation result is classified by the FCN-based DenseNet into 7 different categories. The classification 
accuracy, recall and AUC scores of $98\%$, $98.5\%$, and $99\%$ are achieved on 
HAM10000 dataset of over 10000 images.

\subsection{Summary}
A summary of the CRF methods for pathology image analysis is exhibited in Table~\ref{CRFsum}., which contains some essential attributes of the summarized research papers. Each row indicates publication year, reference, research team, disease, input data, task, inference 
algorithm, classifier, result evaluation. Similar to the MRFs mentioned before, the CRFs is 
also widely applied to various diseases and mostly appears in segmentation and 
classification tasks. The mean-field approximation is the most common way to approximate 
maximum posterior marginal inference, and further explanation will be given in 
Section~\ref{sec:methodan}. Since 2016, DL network is applied on a wide range of computer 
vision tasks, including our research area especially related to the CRFs. Hence, features are rarely extracted in an independent step, because the prepossessed image can be the input 
of DL method, which is more convenient. As shown from the chart, majority of the work achieves an accuracy of over $90\%$.
\begin{table*}
    \centering
    \caption{Summary of reviewed works for pathology image analysis using CRFs. (Matthews correlation coefficient (MCC), Free Response Receiver Operating Characteristic (FROC), intersection over union Coefficient (IoU), Aggregated Jaccard Index (AJI), mean intersection over union (mIOU), overall pixel accuracy (OPA), mean of accuracy (mAC), mean Dice Coef cient (mDC), mean Jaccard Index (mJI), mean thresholded Jaccard index (mTJI), mean Boundary Recall (mBR). )}
    \scriptsize
    \begin{tabular}{p{1.5cm}<{\raggedright}p{1.5cm}<{\raggedright}p{2.5cm}<{\raggedright}p{2cm}<{\raggedright}p{2cm}<{\raggedright}p{1.5cm}<{\raggedright}p{3.5cm}<{\raggedright}}
    \Xhline{1.2pt}
        \textbf{Year, Ref, Research team} & \textbf{Disease} & \textbf{Input data} & \textbf{Task} & \textbf{Inference algorithom} & \textbf{Classfier} & \textbf{Result evaluation} \\ \Xhline{1.2pt}
        2009,\cite{Xu2009Conditional}, Wu et al. & -- & FNA cytological samples from thyroid nodules, 12 images & Segmentation (cell or intercellular material) & Loopy belief propagation & -- & -- \\ \hline
        2010,\cite{park2010semantic}, Park et al. & Cervical cancer & Colposcopy images, 48 patients & Segmentation, classification (normal or abnormal) & -- & KNN, LDA & Sn=70\%, Sp=80\%. \\ \hline
        2011,\cite{park2011domain}, Park et al. & Cervical cancer & Colposcopy images, 48 patients & Segmentation, classification (normal or abnormal) & -- & KNN, LDA & AUC=0.8012. \\ \hline
        2011,\cite{rajapakse2011staging}, Rajapakse et al. & -- & Lung tissue, 9551 cells & Classification (benign cells or cancer cells) & -- & SVM & Acc=90.26\%, Sn=48.30\%, Sp=96.52\%. \\ \hline
        2011,\cite{Xavier2011Brain}, Descombes et al. & Brain tumor & Micro-tomography vascular networks & Segmentation of vascular networks (normal or tumor) & Simulated annealing & -- & -- \\ \hline
        2012,\cite{fu2012glandvision}, Fu et al. & -- & Human colon tissues, 1072 glands & Gland detection and segmentation & Viterbi algorithm & SVR & Acc=70.32\%. \\ \hline
        2014,\cite{fu2014novel}, Fu et al. & -- & Human colon tissues, 1072 glands; 24 H\&E stained images, 333 glands & Gland detection and segmentation & Viterbi algorithm & SVM, SVR  & Dataset 1: Acc=80.4\%; Dataset 2: Sn=82.35\%, Sp=93.08\%, Acc=87.02\%, D=87.53\%. \\ \hline
        2014,\cite{manivannan2014brain}, Manivannan et al. & Brain tumor & Brain tissue, 35 images & Region Segmentation (necrotic regions or normal regions) & Graph cuts & SVM (Bag-of-words) & Acc=66\%. \\ \hline
        2014,\cite{Orlando2014Learning}, Orlando et al. & Eye disease & Fundus retinal images, DRIVE  public datasets &  Retinal vessel segmentation & Mean-field inference & SOSVM & Sn=78.5\%, Sp=96.7\%. \\ \hline
        2016, \cite{wang2016deep}, Wang et al. & IBD & Intestinal tissue, 200 images & Semantic segmentation (muscle or messy regions) & -- & FCN & Acc=90\%, IU=56\%. \\ \hline
        2016, \cite{Fu2016Retinal}, Fu et al. & Eye disease & Fundus retinal images, DRIVE, STARE public datasets &  Retinal vessel segmentation & Mean-field inference & CNN & DRIVE: Acc=94.70\%, Sn=72.94\%; STARE:  Acc=95.45\%, Sn=71.40\%. \\ \hline
        2016,\cite{fu2016deepvessel}, Fu et al. & Eye disease & Fundus retinal images, DRIVE, STARE, and CHASE DB1 public datasets &  Retinal vessel segmentation & Stochastic gradient descent & CNN & DRIVE: Acc=95.23\%, Sn=76.03\%; STARE:  Acc=95.85\%, Sn=74.12\%; CHASE DB1: Acc=94.89\%, Sn=71.30\%. \\ \hline
        2016, \cite{Orlando2016Discriminatively}, Orlando et al. & Eye disease & Fundus retinal images, DRIVE, STARE, CHASEDB1 and HRF public dataset &  Retinal vessel segmentation & SOSVM & -- & DRIVE: Sp=98.02\%, Sn=78.97\%; STARE: Sp=97.38\%, Sn=76.92\%; CHASE DB1:  Sp=97.12\%, Sn=72.77\%; HRF: Sp=96.80\%, Sn=78.74\%. \\ \hline
        2017, \cite{Zhou2017Improving}, Zhou et al. & Eye disease & Fundus retinal images, DRIVE, STARE, CHASEDB1 and HRF public dataset &  Retinal vessel segmentation & Structured support vector machine, fast inference & Modified version of MatConvNet & DRIVE: F1-score= 0.7942, MCC=0.7656,  G-mean=0.8835; STARE: F1-score=0.8017, MCC=0.7830, G-mean=0.8859; CHASEDB1: F1-score=0.7644, MCC=0.7398, G-mean=0.8579; HRF: F1-score=0.7627, MCC=0.7402, G-mean=0.8812.  \\ 
         \Xhline{1.2pt}
    \end{tabular}
    \label{CRFsum}
\end{table*}

\renewcommand\arraystretch{1.5}
\begin{table*}
\scriptsize
    \begin{tabular}{p{1.5cm}<{\raggedright}p{1.5cm}<{\raggedright}p{2.5cm}<{\raggedright}p{2cm}<{\raggedright}p{2cm}<{\raggedright}p{1.5cm}<{\raggedright}p{3.5cm}<{\raggedright}}
   \Xhline{1.2pt}
        \textbf{Year, Ref, Research team} & \textbf{Disease} & \textbf{Input data} & \textbf{Task} & \textbf{Inference algorithom} & \textbf{Classfier} & \textbf{Result evaluation} \\ 
        \Xhline{1.2pt}
         2017, \cite{wu2017STUDY}, Wu et al. & Breast cancer & Breast cancer biopsy images, 1136 images & Mitosis detection & Error back propagation & AlexNet & F=43.7\%. \\ \hline
        2017, \cite{he2017Reasearch}, He et al. & -- & Pluripotent stem cells, 306 sequences & Morphological change pattern classification (healthy, unhealthy or dying) & Margin maximization estimation & LCRF & Acc=93.46\%. \\ \hline
        2018, \cite{li2018cancer}, Li et al. & Breast cancer & WSIs, 400 images & Classification (normal or tumor) & Mean-field inference & ResNet-18 and ResNet-34 & FROC= 0.8096. \\ \hline
        2018, \cite{liang2018Weakly}, Liang et al. & Gastric cancer  & Gastric tumor tissue, 1400 images & Segmentation (normal or tumor) & -- & FCN & IoU=85.51\%. \\ \hline
        2018,\cite{liang2018deep}, Liang et al. & Gastric cancer  & Gastric tumor tissue, 1400 images & Segmentation (normal or tumor) & -- & DU-Net & IoU=88.4\%. \\ \hline
        2018, \cite{Cl2018Multitask}, Playout et al. & Diabetic retinopathy & Fundus retinal images, DIARETDB1 public dataset & Red and bright retinal lesions detection and segmentation & Adadelta algorithm & Modified version of U-Net & Red lesions: Sn= 66.9\%, Sp=99.8\%; bright  lesions: Sn= 75.3\%, Sp=99.9\%. \\ \hline
        2018, \cite{Huang2018Automatic}, Huang et al. & Eye disease & Fundus retinal images, DRIVE, STARE, MESSID, DIARETDB, DIARETD and DRION  dataset & Optic disc identification & Mean-field inference & CNN & DRIVE, MESSID, DIARETDB, and DRION: Acc=100\%; STARE: Acc=98.90\%; DIARETD: Acc=99.90\%. \\ \hline
        2019, \cite{qu2019weakly}, Qu et al. & Lung Cancer  & H\&E stained histopathology images, lung cancer and MultiOrgan dataset & Nuclei segmentation & Mean-field inference & Modified version of U-net & Lung Cancer dataset: Acc=98.1\%, F=92.6\%, D=93.9\%, AJI= 93.2\%; MultiOrgan dataset: Acc=98.7\%, F= 96\% \\ \hline
        2019, \cite{sara2019use}, Jamal et al. & -- & Histopathological images, 58 images & Cellular segmentation & Iterated CRF & SVM, random forest, KNN & F=86.07\%, Kappa=80.28\%, Acc=91.79\%. \\ \hline
        2019, \cite{Zormpas2019Superpixel}, Konstantinos et al. & Melanoma  & Melanoma skin cancer, 105 images & Single-cell classification (cancer cells, lymphocytes, stromal cells or epidermal cells) & Stochastic gradient descent & SC-CNN & Acc=96.48\%, P=96.44\%, R=96.29\%. \\ \hline
        2020, \cite{sun2020hierarchical}\cite{Sun2020GHIS}, Sun et al. & Gastric cancer  & H\&E stained gastric histopathological images, 560 images & Segmentation (normal or tumor) & -- & CNNs & Acc=78.91\%, P=41.58\%, D=46.29\%. \\
 \hline
        2020, \cite{li2020automated}, Li et al. & Prostate cancer  & Prostate cancer tissue
microarray, 1211 images & Gleason grading & -- & ASPP and CNNs & mIOU=77.29\%, OPA=89.51\%. \\       
        \hline
        2020, \cite{bhatkalkar2020improving}, Bhatkalkar et al. & Eye disease  & Fundus retinal private dataset, 300 images & Segmentation of optic disc & -- & DeepLabV3 et al. & DC=0.974. \\
        \hline
        2020, \cite{qiu2020inferring}, Qiu et al. & Pigmented skin lesion & Dermoscopy images, ISIC 2017 and PH2 public dataset & Segmentation of skin lesion & Mean-field inference & DCNNs & mAC=96.20\%, mDC=94.14\%, mJI=89.20\%, mTJI=68.10\%. \\
        \hline
        2020, \cite{adegun2020fcn}, Adegun et al. & Skin cancer & Dermatoscopic images, HAM10000 public dataset & Segmentation and classification of skin lesions & Mean-field inference & FCN-Based DenseNet framework & Acc=98.0\%, F1-score=98.0\%, R=98.5\%. \\                     
        \hline
        2020, \cite{Dong2020GECNN}, Dong et al. & Prostate cancer & H\&E stained WSIs of prostate tissue, 116 images & Segmentation (normal or tumor) & Mean-field inference & Resnet-GE-p4m & Acc=85.1\%, AUC=91.0\%. \\                     
         \Xhline{1.2pt}
    \end{tabular}
\end{table*}
\section{Method Analysis and Discussion}
\label{sec:methodan}

\subsection{Analysis of MRF Methods}
According to the researches on the MRF model applied in our field, it most frequently 
appears in segmentation task, and more specifically, plays two roles in most cases:
\begin{itemize}
\item It serves as a post-processing method to refine initial segmentation results
obtained from classical segmentation methods; 
\item It is incorporated into other segmentation algorithms and they produce results 
together.
\end{itemize}

In the first case, the images are always segmented by some 
popular segmentation methods first to obtain initial labels, such as region growing, 
$k$-means, Bayesian classification, Otsu thresholding et al. However, most of these 
models do not consider contextual constraints, which have significant effect on the 
interpretation of visual information~\cite{li2009markov}. Therefore, the MRF is applied 
based on the initial label using spatial dependencies, producing the final labeling. 
Compared to the second situation, initial segmentation is helpful to avoid the MRF 
algorithm falling into local optimal solutions, but it makes the whole process more 
complex. The papers involved in this article are~\cite{mungle2017mrf, won2004segmenting, 
bioucas2014alternating, su2019cell, monaco2008detection}. In the second case, the MRF is employed in the popular segmentation methods mentioned above, solving the limitation of those assumptions. Hence,
random edges in the area due to noise are much less likely to cause false boundaries and better performance can be achieved. The papers involved in this article are~\cite{xu2010markov,dholey2018combining}.

Parameter estimation and function optimization are essential to MRFs. Some algorithms are adopted for estimating observations from a given distribution. Among them, EM and ICM are the most widely used. The EM algorithm is an iterative algorithm for maximum likelihood estimation in the absence of data.
Because of its simplicity and generality of the associated theory, it is broadly 
applicable these years~\cite{dempster1977maximum}. However, compared to other algorithms, the convergence speed of the EM algorithm is relatively slow. The papers involved in this article 
are~\cite{mungle2017mrf,  bioucas2014alternating, dholey2018combining, su2019cell}. 

\begin{figure}[htbp]
  \centering
  \centerline{\includegraphics[width=0.98\linewidth]{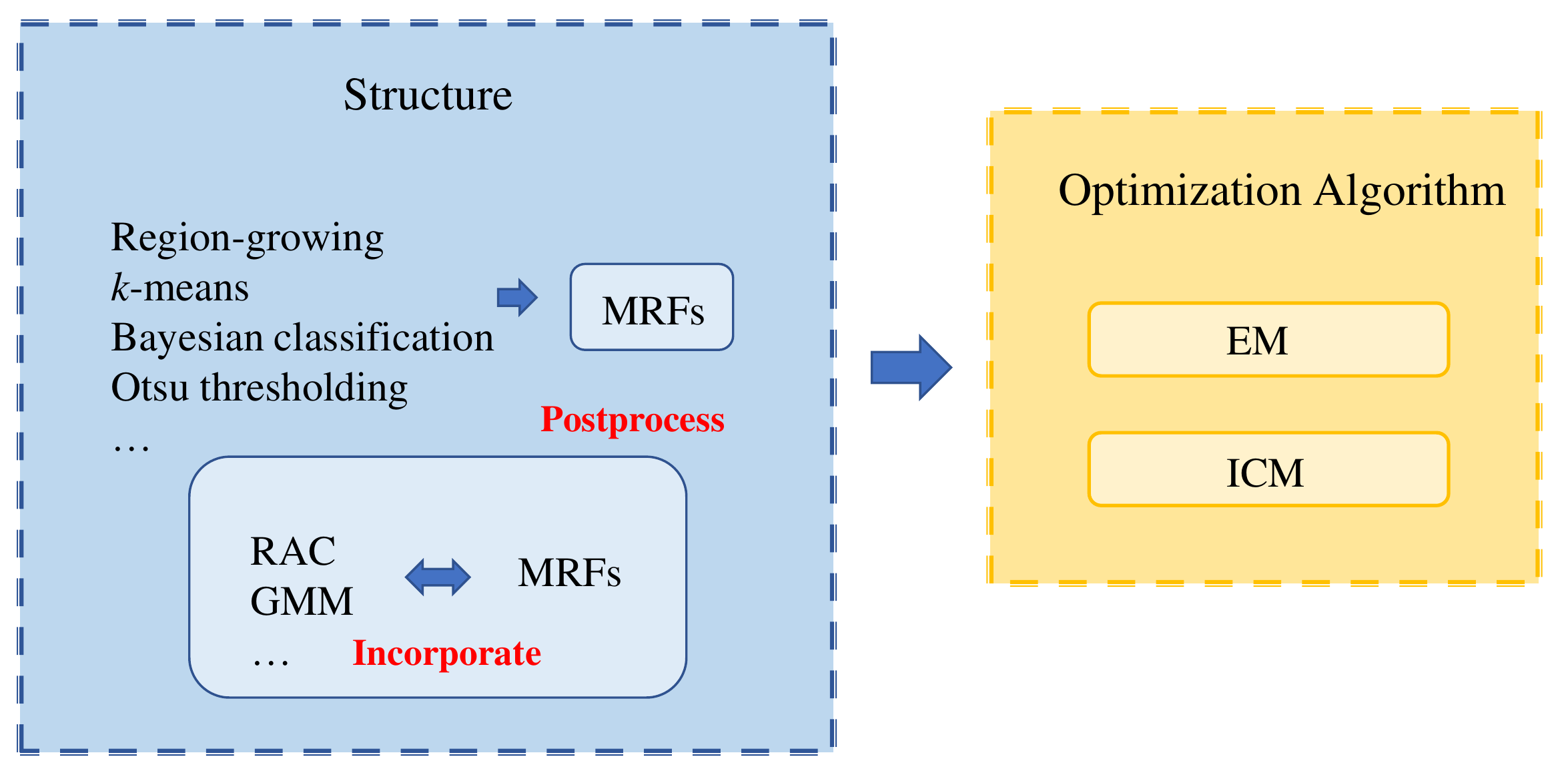}}
\caption{The popular methods in MRFs for pathology image analysis tasks.}
\label{fig:MRFma}
\end{figure}

Compared with the EM algorithm, ICM method not only guarantees the convergence of sequence update, but also has a higher speed, whose computation time can be an order of magnitude less than other commonly used methods. The disadvantage is that different initial states lead to different results, which means that different MAP probabilities not always correspond to meaningful results. It seems to lack of mathematical justification. The papers involved in this article 
are~\cite{mungle2017mrf, basavanhally2009computerized, monaco2008detection,
monaco2010high}. 
In~\cite{monaco2009weighted}, this algorithm is improved to adapt the task requirement, 
which adds a parameter to implicitly weight the importance of each class. The summary 
of the popular methods is shown in Fig.~\ref{fig:MRFma}. 

\subsection{Analysis of CRF Methods}
Nowadays, various improvement measures have been proposed with the increasing influence of CRF. Among them, the FC-CRF which is also called dense CRF, is the most frequently employed discriminative model 
structure in the reviewed tasks, especially for the researches on eye diseases.  Each node of the FC-CRF is a neighbor of each other. Therefore, contextual relationship of different types of labels or long-distance dependencies can be modeled to make the edges smoother and more coherent to semantic objects~\cite{yu2019comprehensive,wu2018image}. The papers involved 
in this article are~\cite{Orlando2014Learning, Fu2016Retinal, fu2016deepvessel, Orlando2016Discriminatively, wu2017STUDY, li2018cancer, liang2018Weakly, qu2019weakly} 
et~\textit{al}. The FC-CRF improves the accuracy, but increases computational cost. To circumvent the limitation, the mean-field approximation, 
a highly efficient approximate inference algorithm for FC-CRF, is proposed and become 
the most common inference approaches in the reviewed papers. The mean-field 
approximation algorithm is used to obtain the marginal label distribution of each 
patch, which is able to achieve significantly more accurate image-labeling performance 
and provide results in less than a second~\cite{krahenbuhl2011efficient}. The 
papers involved in this article are~\cite{li2018cancer, qu2019weakly, Orlando2014Learning, Fu2016Retinal, Huang2018Automatic}.

Driven by the development of machine learning, from 2016, more and more researches 
integrate CRF into DL method, and they can be divided into two categories:
\begin{itemize}
\item The DL network’s outputs form the CRF components, and approximate maximum 
posterior marginal is inferenced by optimization techniques afterwards 
(such as~\cite{Fu2016Retinal, sun2020hierarchical, Zhou2017Improving}); 

\item Firstly, approximate marginal distribution of each patch label is computed 
using the inference algorithm. Then, the CRFs are embedded in the loss function of 
DL network and they are trained together to minimize the loss and achieve optimal 
labels. (e.g.~\cite{li2018cancer, fu2016deepvessel, wang2016deep, Cl2018Multitask}, 
or the CRF loss is used to fine-tune the trained model, e.g.~\cite{qu2019weakly,wu2017STUDY}).\end{itemize}

\begin{figure}[htbp]
  \centering
  \centerline{\includegraphics[width=0.98\linewidth]{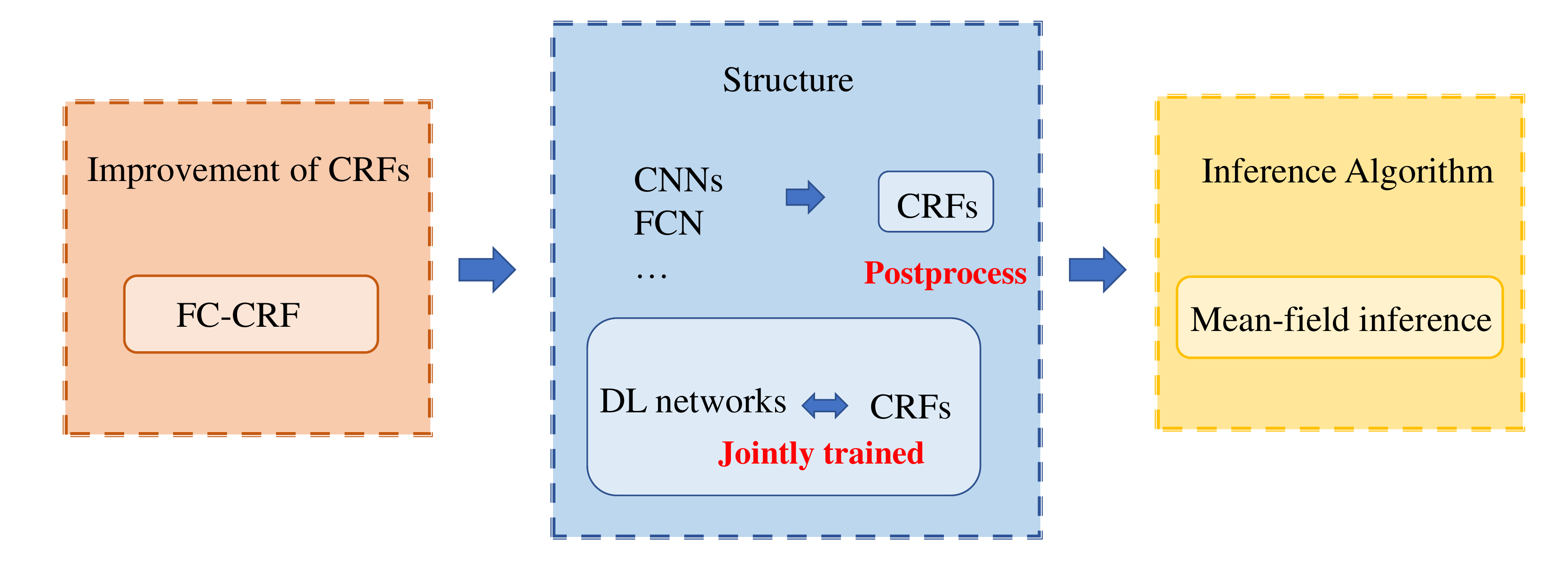}}
\caption{The popular methods in CRFs for pathology image analysis tasks.}
\label{fig:CRFma}
\end{figure}

The former applies the CRFs as a post-processing step after DL network for purpose of incorporating contextual information. However, this does not fully harness the strength of CRFs since it is disconnected from the training of the DL network~\cite{zheng2015conditional}. 
By contrast, the latter combines the DL network and CRF layers into an integrated 
DL architecture, which is an end-to-end learning method that reduces the complexity 
of the project. In~\cite{wang2016deep, fu2016deepvessel, Cl2018Multitask}, the CRFs are implemented as RNNs and embedded in the network as part of CNN, so the back propagation algorithm can be adopted to train the entire deep network end-to-end. The main advantages are as follows: First, DL network's lack of spatial and appearance consistency of the labelling output resulting in poor object delineation and classification accuracy loss. CRFs can be used to overcome this drawback, and this method combines the strengths of both of them. Second, although the DL network can provide a promising result, it has poor interpretability in mathematic theory~\cite{lecun2015What}. However, the DL network and 
CRF are convertible. Therefore the CRF can be used to improve the interpretability of 
DL theory~\cite{zheng2015conditional}. The CRFs have a solid theoretical foundation and 
can be converted into a DL network with good practical results. In conclusion, the CRFs 
have dual advantages in theory and practice, providing a unique perspective for pathology 
image analysis. The summary of the popular methods is shown in Fig.~\ref{fig:CRFma}. 

\subsection{The Potential Random Field Methods in Pathology Image Analysis Field}
In this subsection, we will introduce some possible random field approaches, which have been  applied to other types of image analysis tasks successfully. It can be speculated that they also have potential in pathological image analysis.

\subsubsection{Potential MRF Methods}
With the continuous growth of the datasets, marking annotations associated with various groups of imaging functions and regions of interest is time-consuming and laborious. The max-flow algorithm is frequently applied in the MRF problems. 
In~\cite{iquebal2018unsupervised}, an unsupervised method associated with the max-flow problem is proposed. An MRF prior is taken into consideration before the image's neighborhood structure, and then the flow capacities iteratively gain. Compared to other unsupervised methods, it is greatly improved and reaches more than 90\% dice score in brain tumor MR images segmentation task. In~\cite{koch2017multi}, an approach focusing on solving a multi-atlas segmentation problem is proposed. The study reformulates the issue as a task of finding the optimal solution of MRF energy function. To optimize the objective MRF function, a max-flow-based optimization algorithm is introduced. This method is evaluated on the MR dataset, yielding mean Dice coefficients of nearly 90\%.

The maximum flow technique can use reliable and efficient parallel algorithms for calculation and ensure its convergence, making it meet the requirement for large-scale tag optimization problems~\cite{koch2017multi}. Especially in some whole slide image analysis tasks, it is hopeful that the MRFs with the Max-flow algorithm can handle this kind of work more efficiently.

\subsubsection{Potential CRF Methods}
Rencently, various improvements of the CRFs have been developed and applied in different domain~\cite{yu2019comprehensive}. The multi-label CRF is one of them, which is 
capable to mark one symbol with more than one label by using long-range interactions 
to encode contextual information. 
It can treat non-adjacent markers as one entity, and at the same time detect various types of objects in image segmentation tasks~\cite{wei2015context}. None of the previous researches applies the multi-label CRF for pathology image analysis tasks to the best of our knowledge.

A multi-label CRF for semantic image segmentation task is proposed in~\cite{wei2015context}, where each layer stands for a single object class and is regularized independently. The label space context is modeledby remote interactions between layers, where the sparse inter-layer connections penalize the unlikely occurrence of some groups of labels. They finally achieve state-of-the-art performance on the MSRC-1 and CorelB datasets. 
Moreover, in~\cite{meier2017perturb}, a perturbation-based sampling approach for dense multi-label CRFs is introduced, which is computationally efficient and easy to implement. The method is validated on synthetic and clinical Magnetic Resonance Imaging data, achieving a promising result of the specificity value of 0.88 on the dataset containing 14 patients. 
Furthermore, an unmanned aerial vehicle (UAV) imagery classification problem is formulated 
within a multi-label CRF framework in~\cite{zeggada2018multilabel}. Firstly, an appropriate representation and a classifier with multilayer are used to provide multi-label prediction probabilities in UAV image, which is segmented into patches. Second, the multi-label CRF model is applied to take spatial relationship between adjacent patches as well as labels within the same patch into consideration, aiming to improve the multi-label classification map iteratively. Outstanding performance achieves with 
$83.40\%$ accuracy value.

In pathology image analysis field, most researches divide images into only two class, such as cancer versus non-cancer. However, the categories to be classified are 
not always antithetical. For example, in~\cite{mercan2017multi}, the whole slide breast 
histopathology images can be labelled as five classes. So, it is believed that the multi-label CRF can show outstanding performance 
in the pathology image analysis field, especially in histopathlogy image and colposcopic 
image analysis.

Another improved model of the CRFs is the hierarchical CRF (HCRF) model, which belongs 
to the discriminative models and offers hierarchical and multilevel semantic. The HCRF model inherits the superiority of the CRF model, and achieve the combination of different scales of information~\cite{zhang2015hierarchical} or the short-range as well as the long-range interactions~\cite{kumar2005hierarchical}. 
Some notable applications of the HCRF for image classification problems are introduced below.

In~\cite{zhang2015hierarchical}, an HCRF model is proposed for radar image segmentation. The unary and pairwise potentials are constructed at each scale in order to integrate information in different scales. Besides, the mean-field approximation is employed to inference the best parameters for the HCRF model and maximize the posterior marginal estimation of the model. The proposed method yields a segmentation accuracy value of $89\%$. Similarly, an associative HCRF model is designed in~\cite{yang2018high} to 
improve the classification accuracy of high-resolution remote sensing images. 
The model is constructed on the graphics hierarchy. In this hierarchy, the pixel layer is applied as the base layer, and multiple super-pixel layers are obtained by mean shift pre-segmentation. The proposed model's 
potentials are defined based on pixels' clustered features for super-pixels extracted at 
each layer. This method finally reaches an overall accuracy of $81.59\%$.

It has shown that learning discriminative patterns from the multiscale features delivers 
a more robust classifier with better discriminant performance in pathology image analysis 
task~\cite{jain2020predicting, ning2019multiscale}. It can be because of the way that 
the pathologists diagnose diseases, by examining macroscopical features to microcosmic 
features. These arguments demonstrate the HCRF method's potential, which can be applied 
in our field, especially in histopathology image analysis.

\subsection{The Random Field Methods for Other Potential Fields}
The MRF and CRF methodology discussed in this paper not only can be applied  in pathology image analysis domain but also can perform well in other fields, such as remote 
sensing images, computed tomography images, magnetic resonance imaging, 
and images collected by pipe robots. There are many intelligent diagnosis systems 
developed for high-resolution remote sensing image 
segmentation~\cite{wang2017optimal, troya2018remote}. The remote sensing images 
share the same rotating properties with cytopathology images, which are both without 
directivity. For example, SuperCRF in~\cite{Zormpas2019Superpixel} models cells and super-pixels as nodes as well as edges between the nodes if there is a contextual
connection among nodes by a CRF. Thus, it is highly possible to apply for remote sensing 
image segmentation. Besides, the discussed segmentation algorithms also have potential 
applications in CT and MRI. In microaneurysms (MAs) detection tasks, there are two 
difficulties: nonuniformity of background intensity and the unequal amount of 
background pixels and MAs. Thrombus~\cite{takasugi2017detection, lopez2018fully} 
and pulmonary nodule~\cite{messay2010new} detection tasks also have these 
characteristics. Moreover, the thrombus geometric structure is irregular, which
is similar to that of the optic disk. Thus, the MRF applied 
in~\cite{Razieh2016Microaneurysms} has a possibility to employ in these two fields.
The instability of light in tunnels and the camera’s angle and distance from the 
surface are two serious problems when detecting images captured by pipe robot. 
Meanwhile, the endoscopic image analysis also meets this 
question~\cite{loupos2018autonomous, huang2018deep}. So, the CRF model proposed 
in~\cite{park2010semantic, park2011domain} can find significant use in this area.
In conclusion, MRF and CRF methods summarized in this survey can offer a new perspective to the research in other domains.

\section{Conclusion}
The MRFs and CRFs have attracted significant attention from researchers in various research fields since being proposed. This paper reviews the recent study about the MRF and CRF models applied in pathology image analysis. First, this review 
presents an introduction to random field models and pathology. Second, it elaborates 
on both models’ background knowledge, including their property, modelling, and 
inference processes. Thirdly, the research papers of random fields based pathology image analysis are comprehensively summarized and some prevalent methods are 
discussed, which are grouped according to their tasks and categories of images, respectively. 
Finally, the conclusion are given in this section. The review shows that pathology 
image analysis using the MRFs and CRFs is an increasing topic of interest and improves 
the performance. Besides, they are believed to be applicable to a wider range of research fields to better solve some problems in the future.

%

\begin{acknowledgements}
This work is supported by the ``National Natural Science Foundation of China'' (No. 61806047), 
the ``Fundamental Research Funds for the Central Universities'' (No. N2019003) 
and the ``China Scholarship Council'' (No. 2018GBJ001757). 
We thank Miss Zixian Li and Mr. Guoxian Li for their importantsupport and discussion in this work.
\end{acknowledgements}

%
%

\bibliographystyle{spphys}       
\bibliography{reference}   

%
%

\end{document}